%% file: root.tex
\documentclass[letterpaper, 10 pt, conference]{ieeeconf}

\IEEEoverridecommandlockouts                  

\usepackage{tabto}
\usepackage{url}
\usepackage{graphicx}
\usepackage{svg}
\usepackage{diagbox}
\usepackage{siunitx}
\usepackage[linesnumbered,ruled,vlined]{algorithm2e}
\usepackage{flushend}
\usepackage{amssymb}
\usepackage{amsmath}
\usepackage{import}
\usepackage{subcaption}
\usepackage{makecell}
\usepackage{booktabs}
\usepackage{array}
\usepackage{multirow}
\usepackage{censor}
\DeclareMathOperator*{\argmax}{arg\,max}

\title{\LARGE \bf
Optimized Wrench Polytope Analysis for Real-Time Stability Control of Legged Robots in Complex Multi-Contact Configurations$^{*}$
}

\author{Friedrich Graaf$^{1}$$^{**}$, Elias Birkefeld$^{1}$$^{**}$, Christian Eichmann$^{1}$, Elias Hofele$^{1}$, Tristan Schnell$^{1}$, Georg Heppner$^{1}$,\\Arne Roennau$^{1, 2}$, Rüdiger Dillmann$^{1}$
\thanks{$^{1}$\ FZI Research Center for Information Technology, Karlsruhe, Germany.}%
\thanks{$^{2}$\ Machine Intelligence and Robotics Lab (MaiRo), Karlsruhe Institute for Technology (KIT), Karlsruhe, Germany.}%
\thanks{$^{**}$\ The authors contributed equally to this paper.}%
\thanks{$^{*}$\ This work has been submitted to the IEEE for possible publication.
Copyright may be transferred without notice, after which this version may
no longer be accessible.}%
}

\begin{document}

\maketitle
\thispagestyle{empty}
\pagestyle{empty}

\begin{abstract}
Legged robots offer a variety of automation applications in real-world scenarios.
But areas that are difficult to traverse, like slopes, caves, or scaffolding, still pose a great challenge for traversal.
To tackle this problem, we propose an optimized algorithm for evaluating the full actuatable wrench polytope for arbitrary contact scenarios.
With our improved analysis algorithm, the torques for each joint of the robot can be calculated within a control frequency of $49\ Hz$.
The achieved speedup allows for deployment within a regular control loop for actuating robot poses for different contact scenarios.
We evaluated our stability controller extensively in simulation scenarios and validated its applicability by deploying it on actual walking robot hardware.
The proposed controller achieved stability in very complex scenarios that are currently not achievable by any other controller.

\end{abstract}

\section{INTRODUCTION}
\label{chap:introduction}
While a multitude of robotic systems and stability criteria exist, the majority of walking robots are still bound to simple environments.
Even with current research in robotic movement capabilities, complex environments that include crevices and big obstacles that need to be overcome are still a major problem for traversal. 
Even with the high mobility of legged robots, obstacles that require the robot to climb are beyond current robotic platforms.
Multiple criteria exist, but most of them only operate in 2D or 2.5D space.
However, with the limitation to 2D and 2.5D space, three-dimensional environments like caves and crevices are excluded.
The reason is that current stability criteria do not achieve the real-time performance necessary for control in 3D space.
To solve this issue, we propose an algorithm for efficient evaluation of the full actuatable wrench in 3D space to achieve a real-time applicable torque control. 
Our contribution in this paper can be summarized into three main contributions:
\begin{itemize}
  \item Introduction of a novel analysis approach of the wrench polytope for efficient 6D intersection calculations
  \item Implementation of a pose controller taking advantage of the high-frequency wrench calculation of the analysis
  \item Evaluation of the proposed controller in extensive simulation setups and on the real walking robot LAURON~VI
\end{itemize}

In Section~\ref{chap:related_works}, an overview of previously and currently used algorithms is shown.
Section~\ref{chap:robot_stability} describes the basic structure of the proposed algorithm.
Section~\ref{chap:contact_force_volumes} and Section~\ref{chap:fwp_com} describing the algorithmic details.
In Section~\ref{chap:results_and_experiments}, the tests and evaluations are discussed.

\section{RELATED WORKS}
\label{chap:related_works}

\begin{figure}
    \centering
    \includegraphics[width=\linewidth]{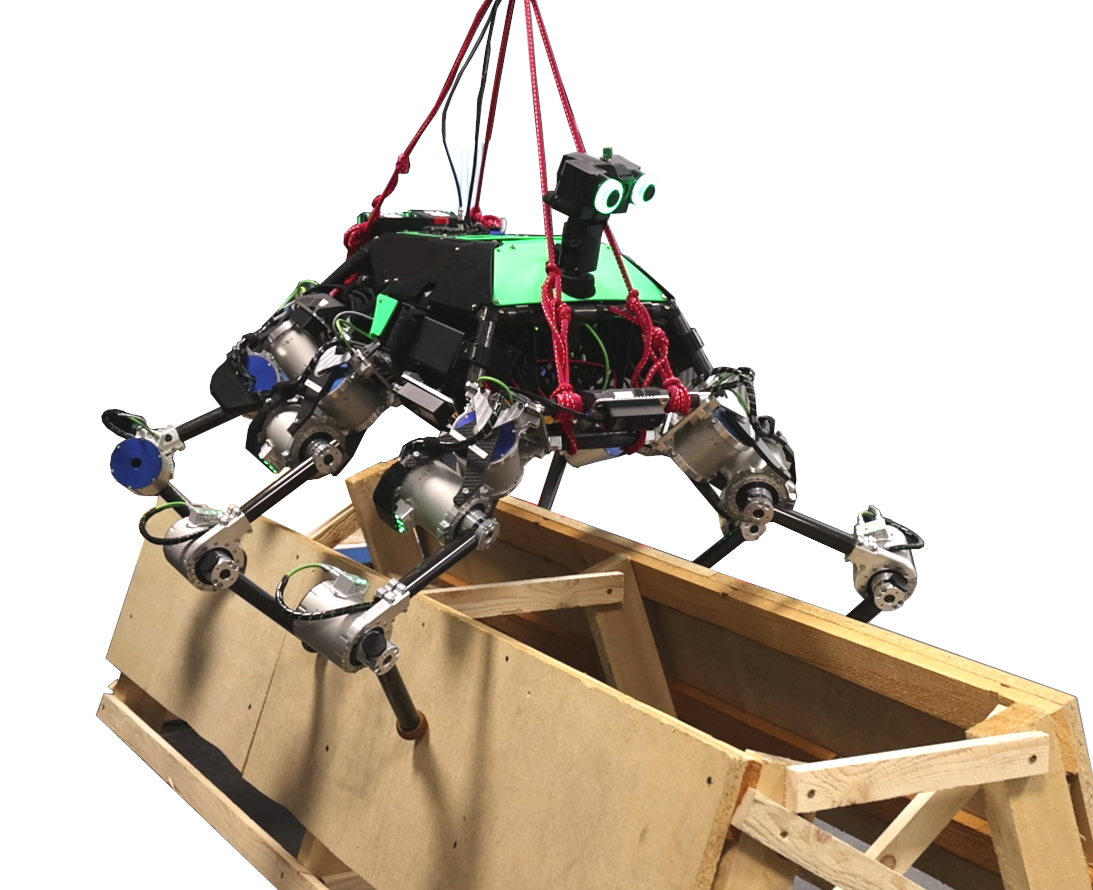}
    \caption{Real-world experimental evaluation of holding onto an angled wall-like structure with LAURON~VI \cite{eichmann_lauron_2025}.
 The ropes were not supporting the weight during the evaluation of the controller.}
    \label{fig:hardware_setup}
\end{figure}

Stability criteria for legged robots have been around for a long time, with the support polygon \cite{mcghee_stability_1968} still being used in recent publications \cite{fankhauser_robust_2018} and the from this derived stability margin \cite{siciliano_springer_2016} being some of the first.
The basic concept is that the center of mass ($CoM$) of the robot has to be within an area in which forces in any direction can be applied.
As this does not fully reflect the physical properties of movement, other criteria have been developed, like the zero-moment point \cite{vukobratovic_zero-moment_2004}.
This criterion also takes the moments and inertia into account and checks if the point at which all moments cancel each other out lies within the contact area of the foot.
Several offshoots of these algorithms have been developed since then \cite{bretl_testing_2008}, \cite{caron_zmp_2017}, which reworked the projection of the stability regions to include friction and 3D contact placement.
A wrench-based criterion using friction cones was proposed in \cite{hirukawa_universal_2006} and implemented in \cite{koyanagi_pattern_2008}.
In further improvements, the actuation limits of the joints were taken into account  \cite{dai_robust_2016} \cite{orsolino_application_2018}.
These algorithms check if the external forces that need to be counteracted lie within the actuatable wrench polytope.
Another approach is able to walk along fixed slopes relying on simplified friction pyramids as well as a support polygon \cite{focchi_high-slope_2017}.
Newest approaches optimize the calculation time for feasible regions within the support polygon \cite{abdalla_efficient_2023} \cite{orsolino_feasible_2020}.

In current robot control strategies relying on reinforcement learning approaches, stability criteria are mostly included implicitly \cite{vogel_robust_2025}, \cite{rudin_parkour_2025}.
Here, the stability is not calculated, but learned through test runs in simulated environments.
Only recently have basic stability criteria been explicitly included in RL training policies \cite{xie_humanoid_2025}.
Before that, stability was learned implicitly through given pose rewards.

As shown, the present algorithms either rely on assumptions to reduce complexity or are not fast enough for actual real-time applicability in control loops.
To not be limited by these factors, our proposed concept is based on the general criterion for calculating feasible wrench polytopes \cite{orsolino_application_2018}.
This makes it rely on only a few assumptions and is applicable to any 3D environment.
But in comparison to \cite{orsolino_application_2018}, which takes over $0.49s$ to calculate a single $4$ contact scenario in similar detail, we achieve an overall $50\ Hz$ control frequency on a $6$ contact scenario for our approach.
This is on par with the control frequency used by current neural networks for locomotion policies \cite{vogel_robust_2025}.

\section{ROBOT STABILITY CONTROL}
\label{chap:robot_stability}
The proposed stability controller design is centered around the feasible wrench polytope ($FW$).
The $FW$ is a 6D volume that contains all possible forces and torques that the robot can exert on its own center of mass ($CoM$) via its contacts with the environment.
\begin{equation}
 FW \subseteq \mathbb{R}^6
\end{equation}
Given this polytope, external forces like gravity can be counteracted by generating a counterwrench that is contained in this polytope.
This directly leads to the stability criterion, which states that a robot is not stable if there exists a direction in which no wrench can be exerted.
\begin{equation}
    \label{eq:stability}
    \forall \mathbf{d} \in \mathbb{R}^6, \quad \exists \varepsilon > 0 : \mathbf{d} \cdot \varepsilon \in FW
\end{equation}
As formulated in Equation~(\ref{eq:stability}), it requires that any wrench $\mathbf{d}$ is present in the $FW$ with a scaling $\varepsilon$ greater than $0$.
To check if these conditions are fulfilled, the $FW$ needs to be calculated.
For a legged robot, the $FW$ is generated by combining the contact wrench polytopes $CW_c$ of each foot contact $c$.
The $CW_c$ contains all wrenches it can exert on the environment over the contact.
\begin{equation}
  \label{eq:fwp_minkowski}
 FW = \bigoplus_{c=1}^k CW_c
\end{equation}
This means the $FW$ is the Minkowski sum of all $CW_c$ as shown in Equation~(\ref{eq:fwp_minkowski}) with $k$ being the number of contact points.
We chose the Minkowski sum, or vector sum \cite{de_berg_computational_2008}, as it represents a lossless combination of all available wrenches.
It is the sum of two or more spaces representing the polytope of all possible vector combinations of each summand.
The contribution of our proposed controller is the efficient calculation of only a small segment of the $FW$ that contains the necessary information to check if a given wrench is contained.
This is done by combining extreme points of the $CW_c$ and only calculating single points in $FW$.
From these points, new extreme points are calculated that are closer to the intersection point to be found in the hull of $FW$.
The creation of these $CW_c$ will be discussed in detail in Section \ref{chap:contact_force_volumes} while the iterative exploration over the extreme points in $FW$ is shown in Section \ref{chap:fwp_com}.

\section{CONTACT WRENCH POLYTOPE CREATION}
\label{chap:contact_force_volumes}
The 6D $CW_c$ that defines all possible force interactions for one contact point $c$ is defined by the intersection of two underlying polytopes.
These are the friction polytope ($FP_c$) and the actuatable polytope ($AP_c$).
\begin{equation}
 CW_c=FP_c \cap AP_c 
\end{equation}
The $FP_c$ is defined by the contact transmittable wrenches given through the friction model.
Staying within the borders of the $FP_c$ prevents slippage or other loss of contact.
The $AP_c$, on the other hand, is defined by the joint torque limits of the robot's joints.
This emanates from the problem that not every wrench can be applied by every joint configuration of the leg.

\subsection{Actuatable Polytope}
For the generation of the $AP_c$, the joint torque limitations can be used directly to generate a half-space representation ($H$-representation).
\begin{equation}
    \label{eq:h_rep}
 AP_c = \left\lbrace 
 f \in \mathbb{R}^6: 
        \begin{Bmatrix}
 T_{ID} \cdot \mathbf{w} \geq \boldsymbol{\tau}_{min} \\
 T_{ID} \cdot \mathbf{w} \leq \boldsymbol{\tau}_{max}
        \end{Bmatrix}
 \right\rbrace
\end{equation}
The $H$-representation is defined by concatenating multiple half-spaces, which are defined by inequality equations, as shown for the $AP_c$ in Equation~(\ref{eq:h_rep}).
Here $T_{ID}$ represents the inverse dynamics transform, mapping the wrenches $\mathbf{w}$ onto the joint torques $\boldsymbol{\tau}$.
Using the inverse dynamics transform, any wrenches can be transformed into joint space and directly checked if they exceed joint limits.
To get the inverse dynamics transform, the recursive-newton-euler algorithm (RNEA) \cite{siciliano_springer_2016} can be used.
However, the inverse dynamics transform still does not fully display the wrench limitations, as singularities that can not be actuated are still included.
These singularities appear when a joint carries a load parallel to the axis of the joint's rotation.
This would make the kinematic chain transmit forces that are not modeled through the joint movement.
To prevent this behavior, a virtual joint is added in this direction.
These joints have their torque limits set to $0$.
This way, any force that would be transmitted passively through the joint is marked as not actuatable.
Applying this filtering, the actuatable polytope $AP_c$ is fully defined.

\subsection{Friction Polytope}
The $FP_c$ is defined in a vertex representation ($V$-representation) \cite{ziegler_lectures_2012}.
\begin{equation}
FP_c = \left\{ \sum_{i=1}^n \lambda_i \mathbf{v}_i \;\middle|\;
 \mathbf{v}_i \in V_c,\; \lambda_i \geq 0,\; \sum_{i=1}^n \lambda_i = 1 \right\}
\label{eq:v_representation}
\end{equation}
Here, the $FP_c$ is defined as the set of linear combinations of its border vertices $V_c=\{\mathbf{v}_1,\mathbf{v}_2, ..., \mathbf{v}_n\}$ as shown in Equation~(\ref{eq:v_representation}).
These border vertices $V_c$ can be defined by a friction model or measured directly in a test setup.
This is done by applying forces in each direction and checking at what force threshold the contact slips.
Given this measured contact, special foot geometries can be integrated that do not follow the default friction model. 

\begin{figure}
    \centering
    \vspace{4pt}
    \begin{subfigure}{70pt}
        \centering
        \includegraphics[width=\linewidth]{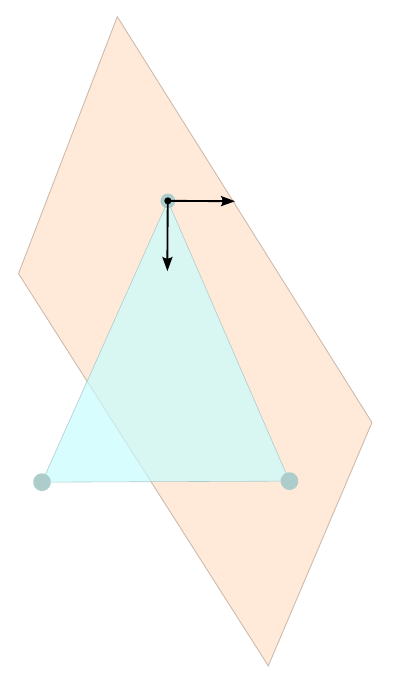}
        \caption{}
        \label{fig:polytope_cut_1}
    \end{subfigure}
    \begin{subfigure}{70pt}
        \centering
        \includegraphics[width=\linewidth]{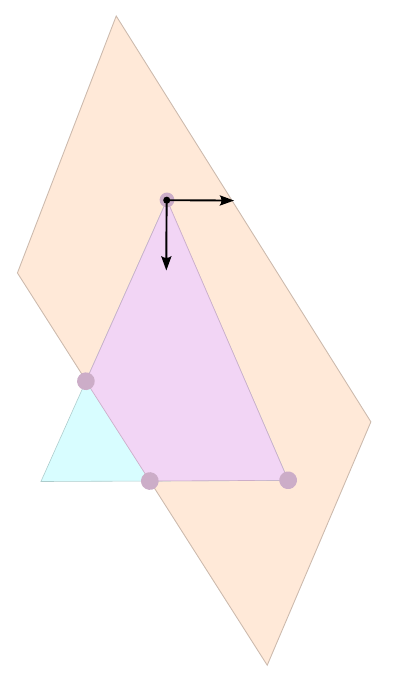}
        \caption{}
        \label{fig:polytope_cut_2}
    \end{subfigure}
    \begin{subfigure}{70pt}
        \centering
        \includegraphics[width=\linewidth]{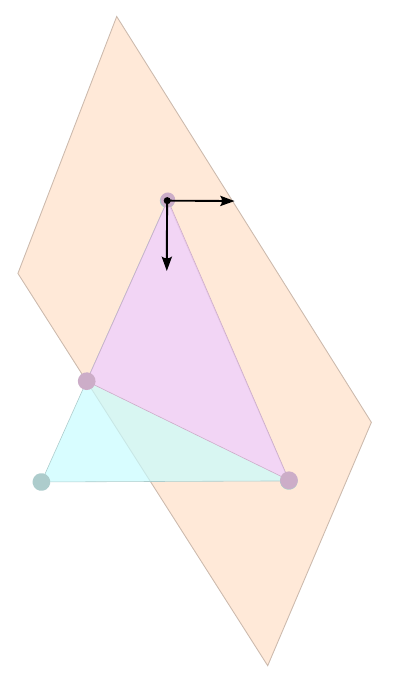}
        \caption{}
        \label{fig:polytope_cut_3}
    \end{subfigure}

    \caption{Simplified example of the contact polytope cut.
 Bright blue is the vertex friction polytope $FP_c$, and beige is the actuatable boundary volume $AP_c$ defined by the joints.
In (a), the original volumes are shown.
 The exact intersection of both volumes is depicted in purple in (b).
 In (c), the result of the scaling intersection method is shown.}
    \label{fig:polytope_cut}
\end{figure}

\subsection{Contact Wrench Polytope}
To combine the $AP_c$ and the $FP_c$ into the actual $CW_c$, they need to be intersected.
The intersection of these two is done by a combination of a precise intersection algorithm and a fast scaling algorithm.

The precise algorithm is the exact intersection of both polytopes by combining their $H$-representations.
This is used for the first contact of each foot, to ensure an exact limitation of friction and actuation limits.
The intersection is shown in Fig.~\ref{fig:polytope_cut_2}.
To generate this intersection, the $V$-representation of the $FP_c$ is converted into an $H$-representation as both are interchangeable \cite{ziegler_lectures_2012}.
The half-space equations are then concatenated to define the $CW_c$ in $H$-representation.
This one is then again transformed back into the $V$-representation.
The backtransformation is necessary, as the latter algorithms depend on a quick evaluation of extreme points, which is faster in $V$-representation.

The faster algorithm of handling this intersection is a simple scaling approach, as shown in Fig.~\ref{fig:polytope_cut_3}.
Here, the $V$-representation of the $FP_c$ is taken directly, and the vertices are checked for the boundary conditions of the $AP_c$ by applying the inverse dynamics transform.
Any vertex $\mathbf{v}_i$ that requires a joint torque $\boldsymbol{\tau}_i$ that exceeds the joint torque limitations is scaled down proportionally to $\mathbf{\tilde{v}}_i$.
\begin{equation}
 T_{ID} \cdot \mathbf{v}_i=\boldsymbol{\tau}_i
  \label{eq:torque_calculation}
\end{equation}
\begin{equation}
 \mathbf{\tilde{v}}_i= \mathbf{v}_i \cdot \min\!\left(1, \min_j \frac{\boldsymbol{\tau}_{max j}}{\boldsymbol{\tau}_{j}}\right)
  \label{eq:scaling}
\end{equation}
The scaling as shown in Equation~(\ref{eq:scaling}) sets the overshoot on the intersection border of both polytopes.
While the calculation of the joint torques $\boldsymbol{\tau}$ is done as described in Equation~(\ref{eq:torque_calculation}), the scaling is shown in Equation~(\ref{eq:scaling}).
For clarity, we omitted the double limit of $\boldsymbol{\tau}_{max}$ and $\boldsymbol{\tau}_{min}$ in the equation.
This results in only applying a simple scaling to build the complete $CW_c$, reducing the calculation time in comparison to the precise approach.
But as shown in Fig.~\ref{fig:polytope_cut}, this also incurs an error, as actual corner points disappear.
In a 6D polytope, this simplification can lead to a loss of up to 80\% of the $CW_c$, which would make the fast approach alone infeasible.

With these two algorithms for intersection defined, we implemented a combination of both to increase calculation speed as well as maintain accuracy.
This was done by executing the exact intersection each time the contact configuration changed and then scaling this intersection's $V$-representation to adapt it for each control loop.
As the $AP$ is mostly defined by the contact configuration and only slightly changes during execution, this method reduces the error to a minimum, while allowing for a high control frequency.

\subsection{Augmenting the Contact Wrench Polytope with CoM Torque Relations}
The wrench polytopes $CW_c$ can still not be used directly to generate the valid $FW$ because the torque they apply to the robots' $CoM$ has not yet been calculated.
The torques are calculated from the force component and the contact's displacement to the robot's $CoM$, resulting in the modified $CWm_c$ polytope.
The displacement vector $\mathbf{r}_c$ is defined as the vector from the $CoM$ to the contact point.
\begin{equation}
CWm_c = \left\{
\begin{bmatrix}
\mathbf{f} \\[4pt]
\mathbf{r}_c \times \mathbf{f}
\end{bmatrix}
\,\middle|\,
[\mathbf{f}, \boldsymbol{\tau}] \in CW_c
\right\}.
\label{eq:cw_torque_calculation}
\end{equation}
The torques $\boldsymbol{\tau}$ from $CW_c$ are initialized as zero, as the $CoM$ position is unknown at that time, which is why they are not used for the calculation of $CWm_c$ in Equation~(\ref{eq:cw_torque_calculation}).

\section{FEASIBLE WRENCH POLYTOPE ANALYSIS}
\label{chap:fwp_com}
After creating the contact wrench polytopes $CWm_c$ for all foot contact points, they must be combined into one feasible wrench polytope $FW$ acting on the robot's $CoM$.
However, the problem is that the calculation of the full Minkowski sum for this combination is not feasible for real-time applications.
As the Minkowski sum of $FW$ is a combination of all possible foot wrenches in $CWm_c$, the number of vertices and interconnections between them grows factorial.
Given a robot with six legs, the border of the sum easily reaches up to $10000$ vertices, with facelets far above that.

To avoid working with this huge amount of data for each time step, we propose data-efficient algorithms similar to the Johnson-Gilbert-Kireeti algorithm \cite{gilbert_fast_1988} to reduce the calculation load of the process.
This enables the main stability algorithm to calculate the individual foot wrenches through the interaction with only parts of the Minkowski sum.
The proposed algorithmic steps are explained in detail below.

\subsection{Initialization of the Iterative Algorithm}

\begin{figure}
    \centering
    \vspace{4pt}
      \begin{subfigure}{0.49\columnwidth}
        \def\svgwidth{\linewidth}
        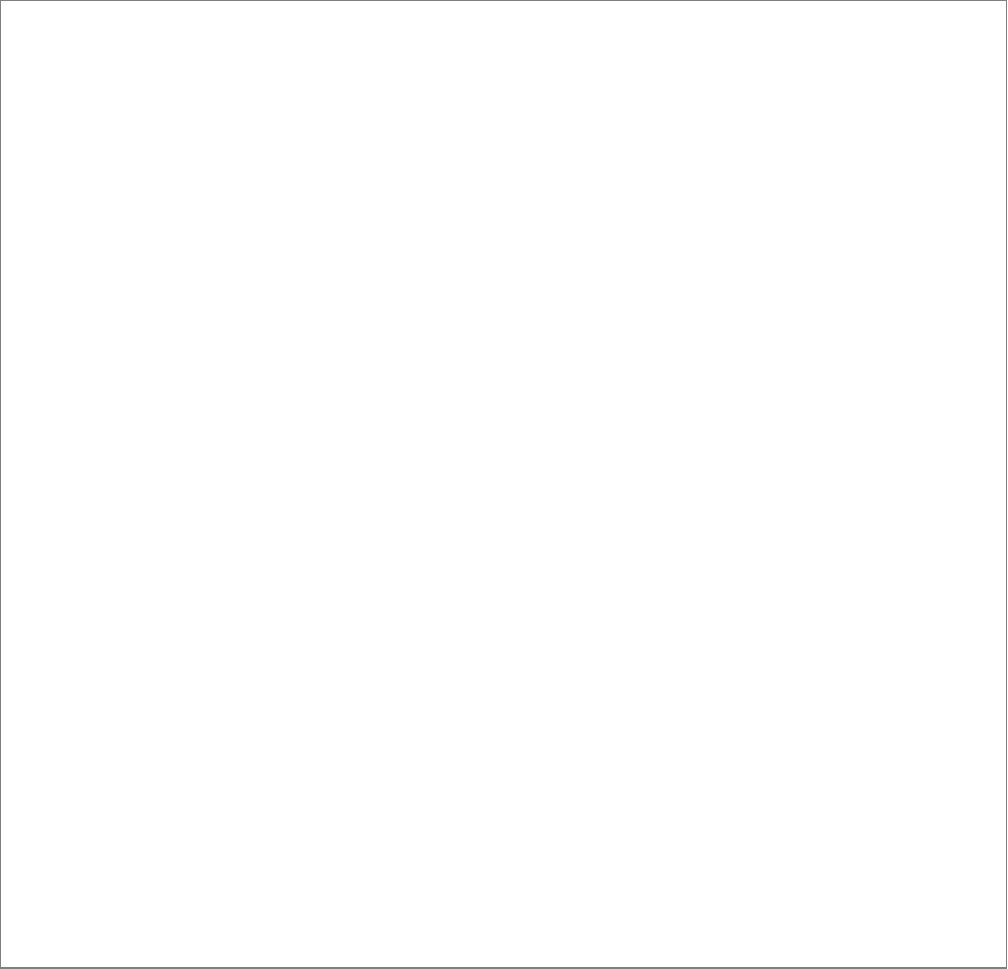
        \caption{Origin with the initial simplex not containing the origin.}
        \label{subfig:initial_simplex_1}
      \end{subfigure}
      \begin{subfigure}{0.49\columnwidth}
        \def\svgwidth{\linewidth}
        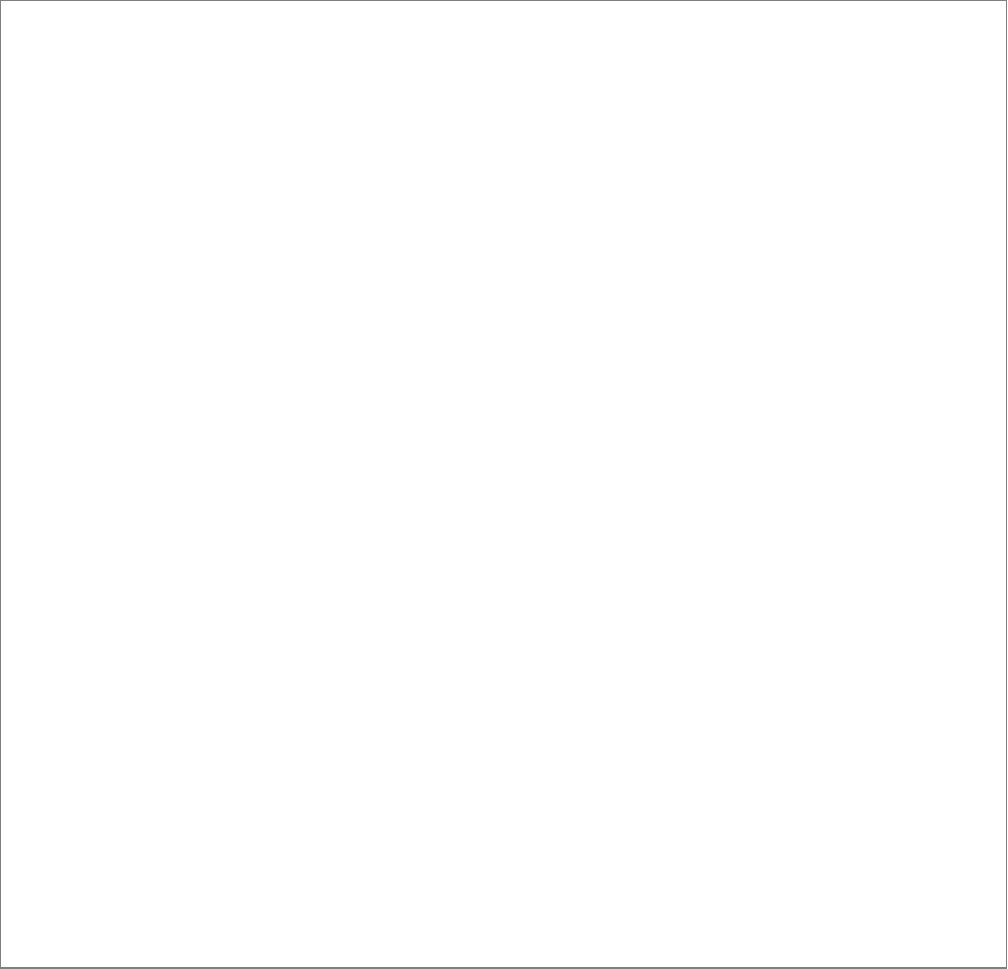
        \caption{Selection of the normals pointing towards the origin.}
        \label{subfig:initial_simplex_2}
      \end{subfigure}

      \begin{subfigure}{0.49\columnwidth}
        \def\svgwidth{\linewidth}
        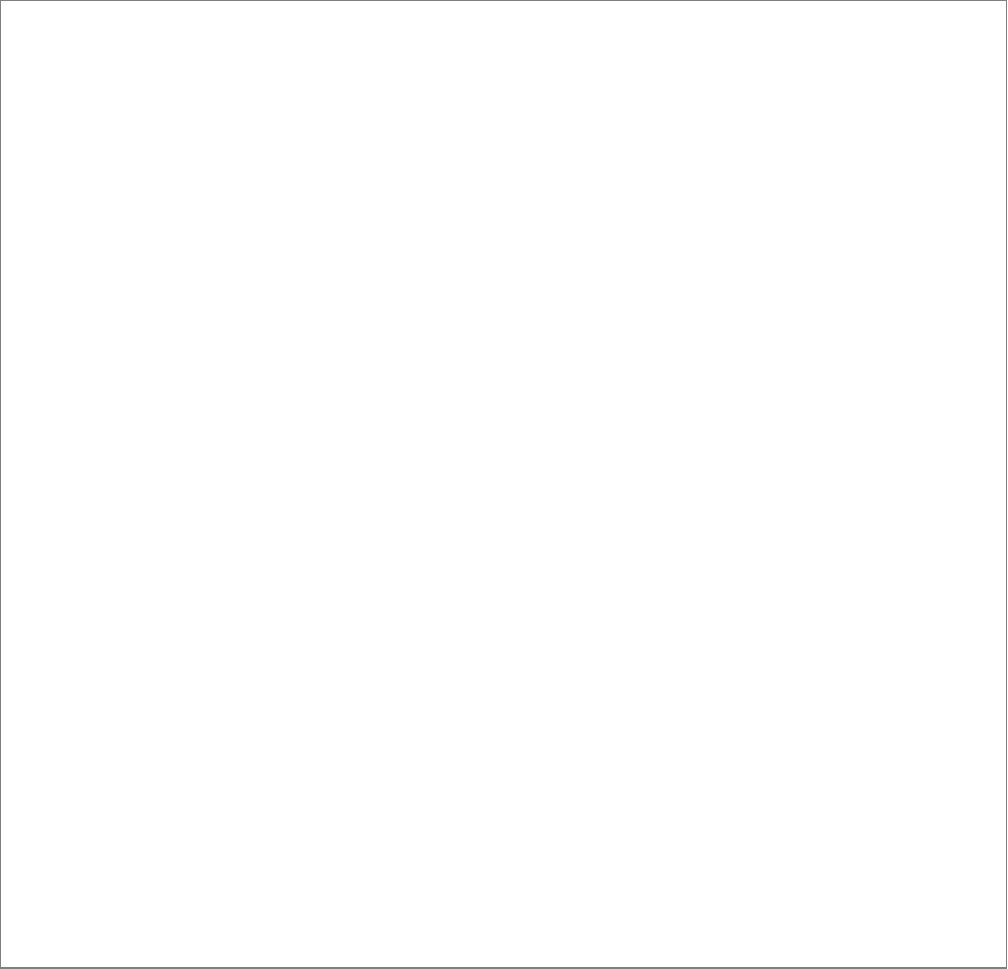
        \caption{Extreme point search in the normal's direction.}
        \label{subfig:initial_simplex_3}
      \end{subfigure}
      \begin{subfigure}{0.49\columnwidth}
        \def\svgwidth{\linewidth}
        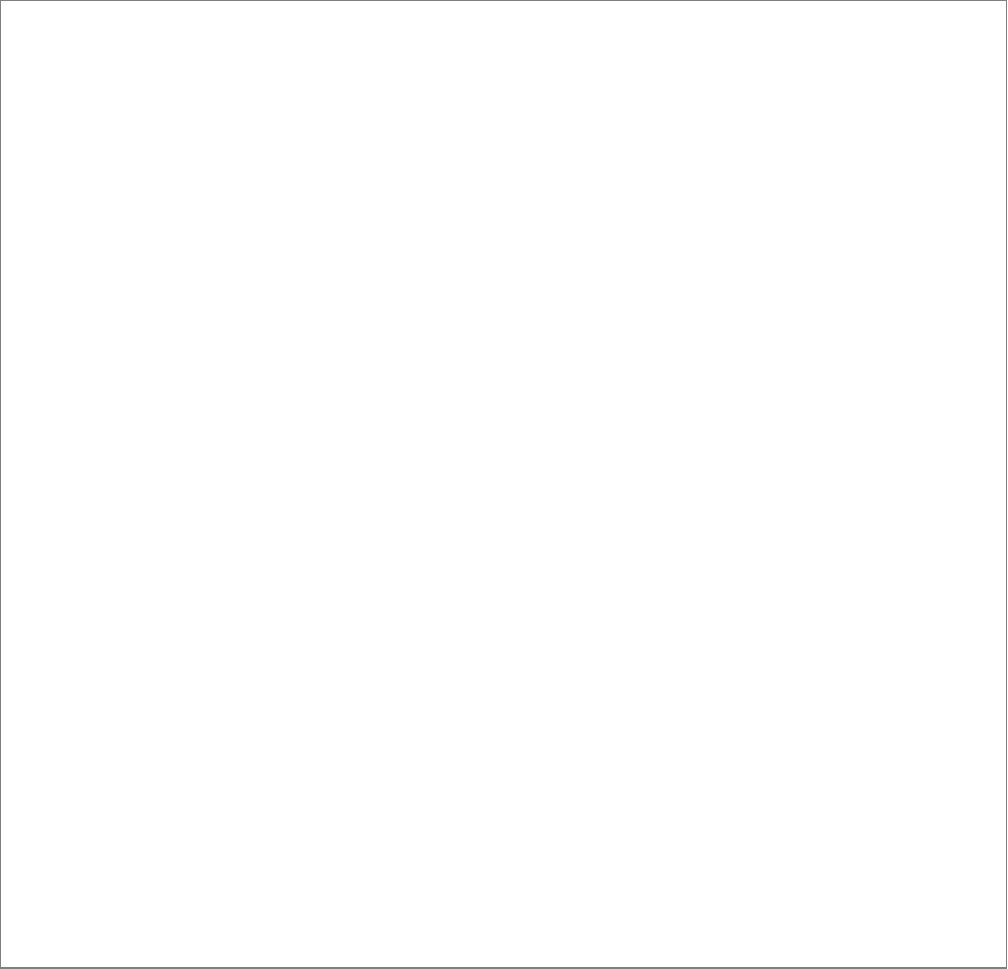
        \caption{Replacement of the simplex point with the new extreme point.}
        \label{subfig:initial_simplex_4}
      \end{subfigure}
      
    \caption{Algorithm to iteratively extend a simplex in the direction of the origin. The process of a single iteration step is shown in subfigures (a) to (d). The algorithm terminates as soon as the origin is contained in the simplex.}
    \label{fig:origin_extension}
\end{figure}

For the initialization, the algorithm requires a goal wrench $\mathbf{w}_{goal}$ to be applied to the robot's $CoM$ as input.
Given this $\mathbf{w}_{goal}$, the wrenches in $CWm_c$ necessary to achieve it need to be determined.
Each wrench $FW_i \in FW$ is saved with its corresponding $CWm_c$ values.
\begin{equation}
  \label{eq:fw_consists_of_cmw}
  FW_i=\{ CWm_1, CMw_2, ..., CMw_c \}
\end{equation}
Because $CWm_c$ has discrete vertices and subsequently $FW$ as well, an exact match with $\mathbf{w}_{goal}$ is unlikely and instead needs interpolation.  
The interpolation will always result in a valid wrench for each foot, as the convexity of the $CWm_c$ polytopes transfers over to the $FW$ through the Minkowski sum.
This also shows that not the whole Minkowski sum needs to be known for the foot wrench calculation, but only the bordering wrenches from which to interpolate.

To find these wrenches without having to calculate the full Minkowski sum, we propose a novel algorithm that uses extreme point searches through the subvolumes.
In our case, we apply it to find a specific section of the outer hull of the Minkowski sum volume where the direction of $\mathbf{w}_{goal}$ intersects.

\begin{figure}
    \centering
    \vspace{4pt}
    \includegraphics[height=150pt]{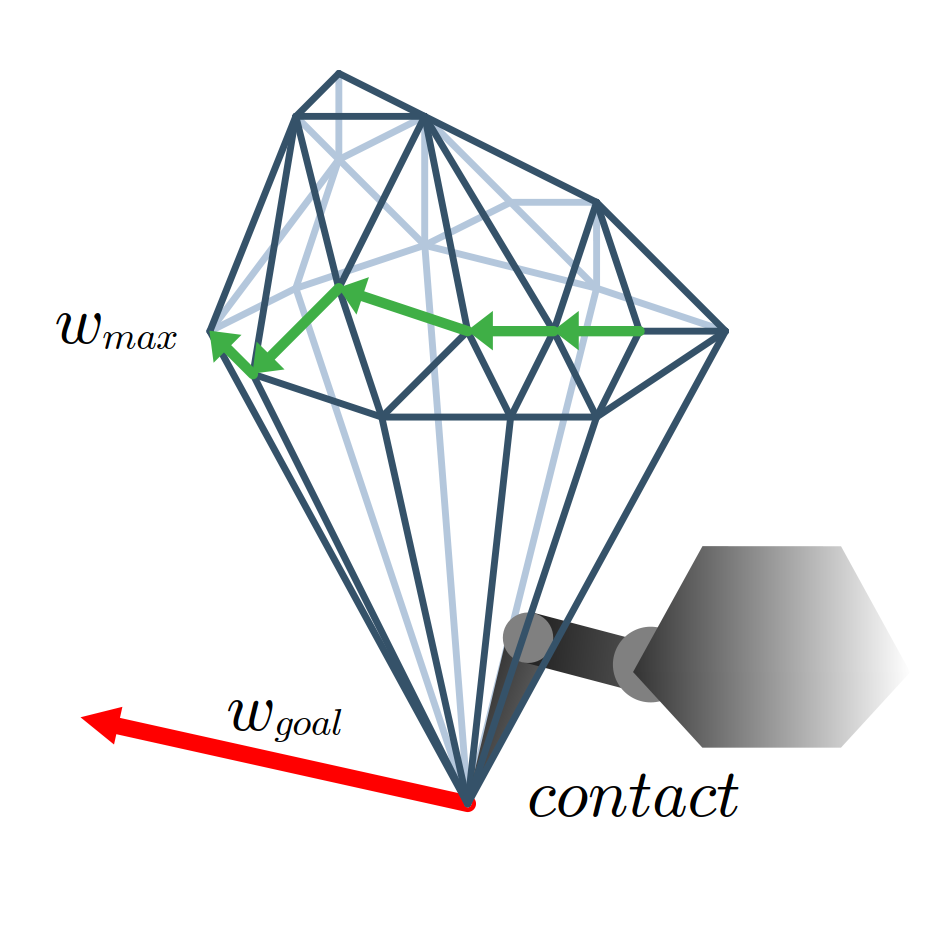}
    \caption{The iteration along the convex hull towards an extreme point of a single $CWm_c$.}
    \label{fig:extreme_point_walk}
\end{figure}

We present an algorithm for finding this intersection efficiently by starting with an initial simplex of the wrench space and extending it towards the outer hull.
A simplex in the context of polytopes is a polytope that contains the minimal number of points to fully enclose a space.
Our extension algorithm needs an initial simplex that contains the origin, so that an intersection of the direction of $\mathbf{w}_{goal}$ with the simplex is ensured.

\subsection{Search for the Origin Simplex}

The generation of the initial simplex is shown in Fig.~\ref{fig:origin_extension}.
First, a random simplex of the Minkowski sum is taken and then extended towards the origin of the wrench space (Fig.~\ref{subfig:initial_simplex_1}).
The basic concept of the extension algorithm starts with selecting the facet of the random simplex that is pointing towards the origin (Fig.~\ref{subfig:initial_simplex_2}).
Along the facet's normal direction $\mathbf{n}$ the extreme point $\mathbf{w}_{max}$ is searched in $FW$ (Fig.~\ref{subfig:initial_simplex_3}).
This extreme point is the sum of the extreme points $\mathbf{c}_{max}$ for each $CMw_c$.
The extreme point $\mathbf{c}_{max}$ is defined as the foot wrench vertex with the highest dot product with $\mathbf{n}$, as shown in Equation~(\ref{eq:extreme_point}).
\begin{equation}
    \label{eq:extreme_point}
\mathtt{extreme\_point}(\mathbf{n}) = \argmax_{\mathbf{c} \in CWm_c} \mathbf{c}^{T} \mathbf{n} = \mathbf{c}_{max}
\end{equation}
The selection of the $\mathbf{c}_{max}$ is not done by calculating the dot product of all points, but instead uses an iterative approach that "walks" along the hull.
This "walk" is done by taking the current wrench and evaluating all neighboring wrench vertices with the dot product.
By doing this, the dot product only has to be executed a few times and not on each full $CWm_c$.
A visualization of this "walk" is shown in Fig.~\ref{fig:extreme_point_walk}.

Given $\mathbf{w}_{max} = \sum \mathbf{c}_{max}$, the new wrench $\mathbf{w}_{max}$ is then used to extend the simplex towards the origin.
This is done by replacing the wrench that had not been part of the facet pointing towards the origin (Fig.~\ref{subfig:initial_simplex_4}).

The $V$-representation has been chosen for the reason of being able to execute this algorithm efficiently.
This way of finding extreme points achieves a speedup by a factor of $10$ times in comparison to the $H$-representation on the same hardware.
For better calculation efficiency, an initial simplex that already contains or is close to containing the origin is preferable.
The reuse of previously calculated origin simplices also greatly reduces this initial calculation time.

\begin{figure}
    \centering
    \vspace{4pt}
    \begin{subfigure}{0.49\columnwidth}
        \def\svgwidth{\linewidth}
        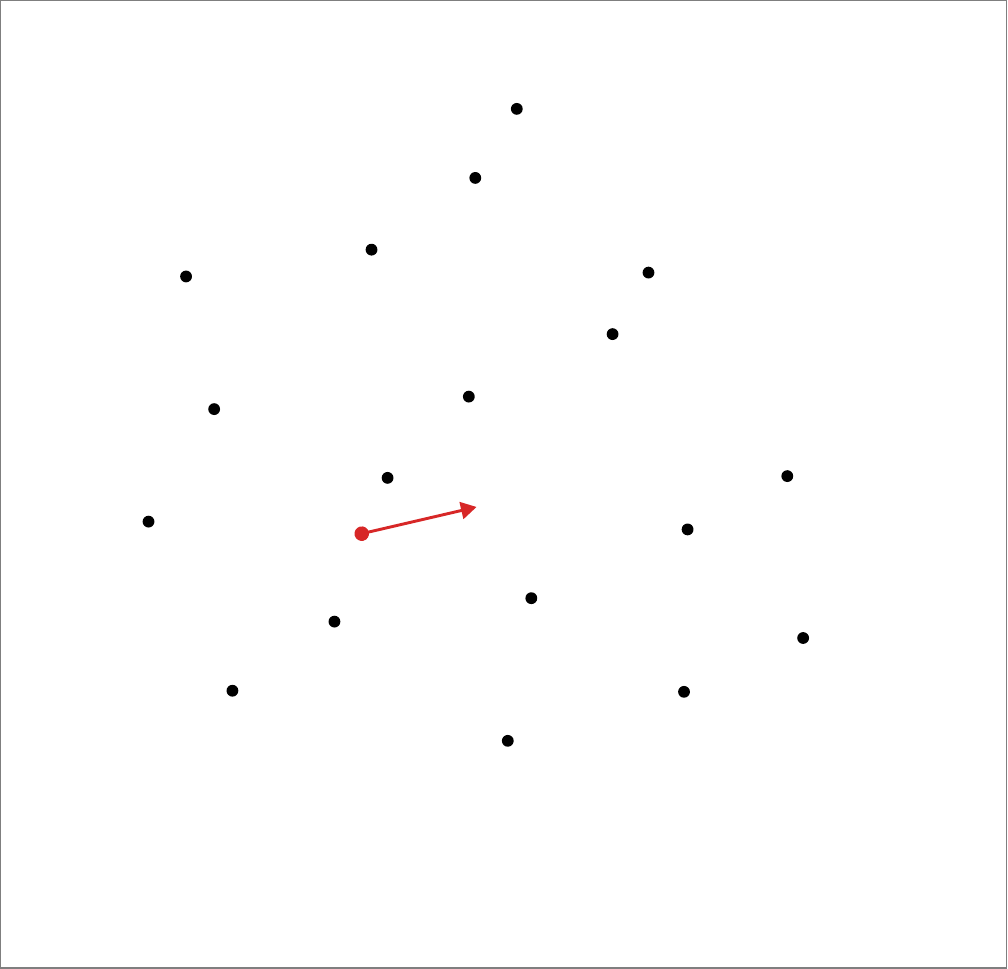
        \caption{Origin $origin$ and goal wrench $goal_{w}$}
        \label{subfig:extend_simplex_1}
      \end{subfigure}
      \begin{subfigure}{0.49\columnwidth}
        \def\svgwidth{\linewidth}
        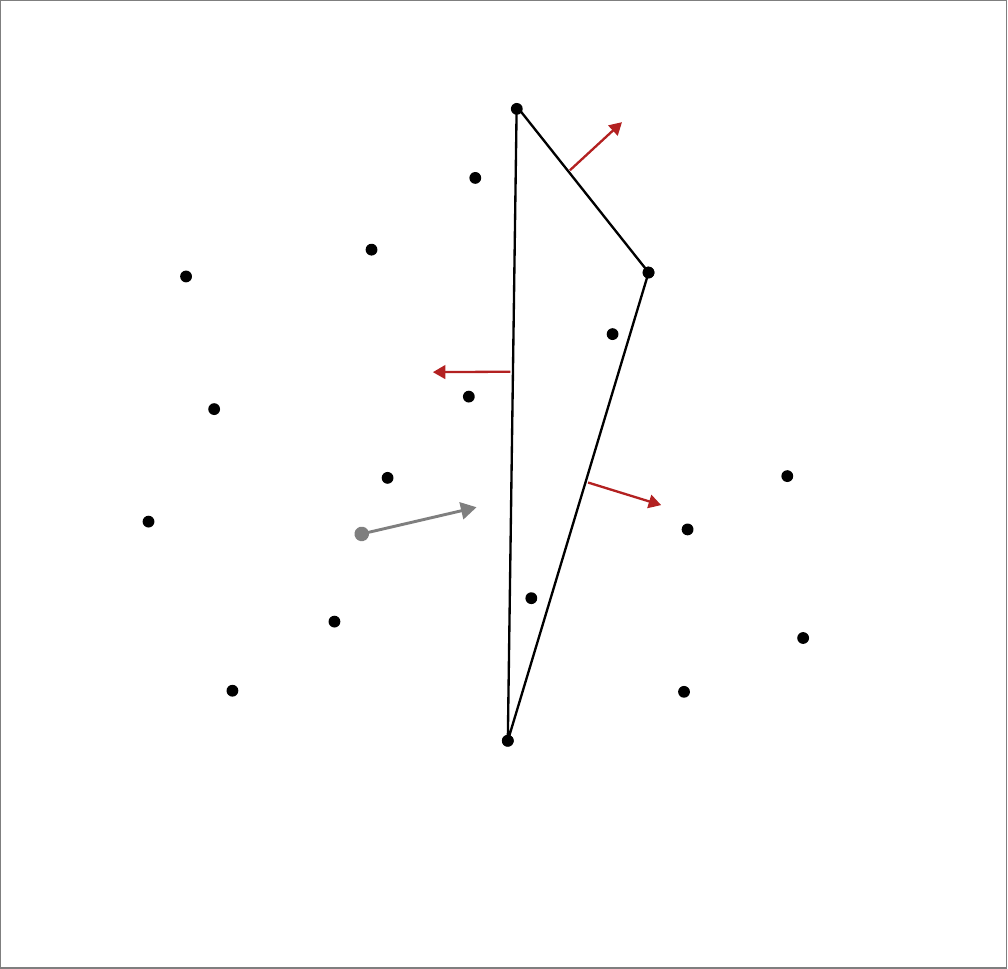
        \caption{Normal orientations of each facet of the current simplex.}
        \label{subfig:extend_simplex_2}
      \end{subfigure}

      \begin{subfigure}{0.49\columnwidth}
        \def\svgwidth{\linewidth}
        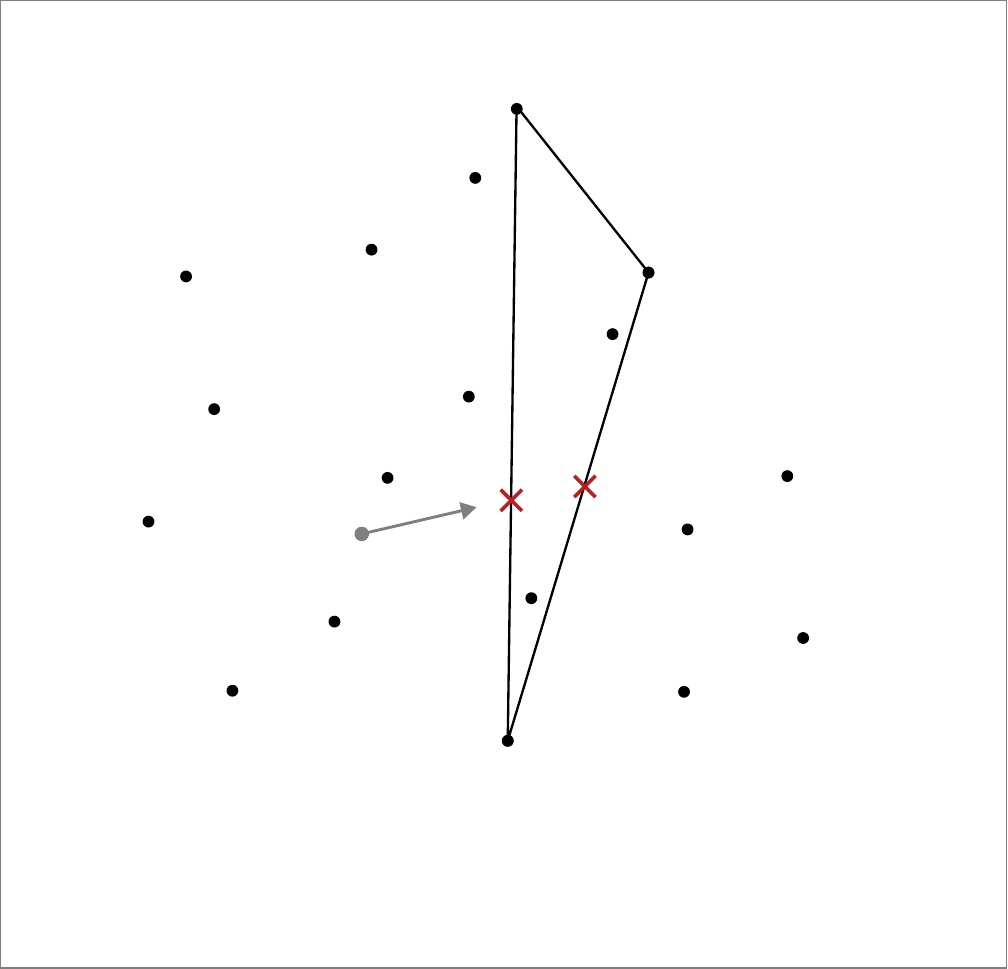
        \caption{Intersection points of the goal wrench with the facets.}
        \label{subfig:extend_simplex_3}
      \end{subfigure}
      \begin{subfigure}{0.49\columnwidth}
        \def\svgwidth{\linewidth}
        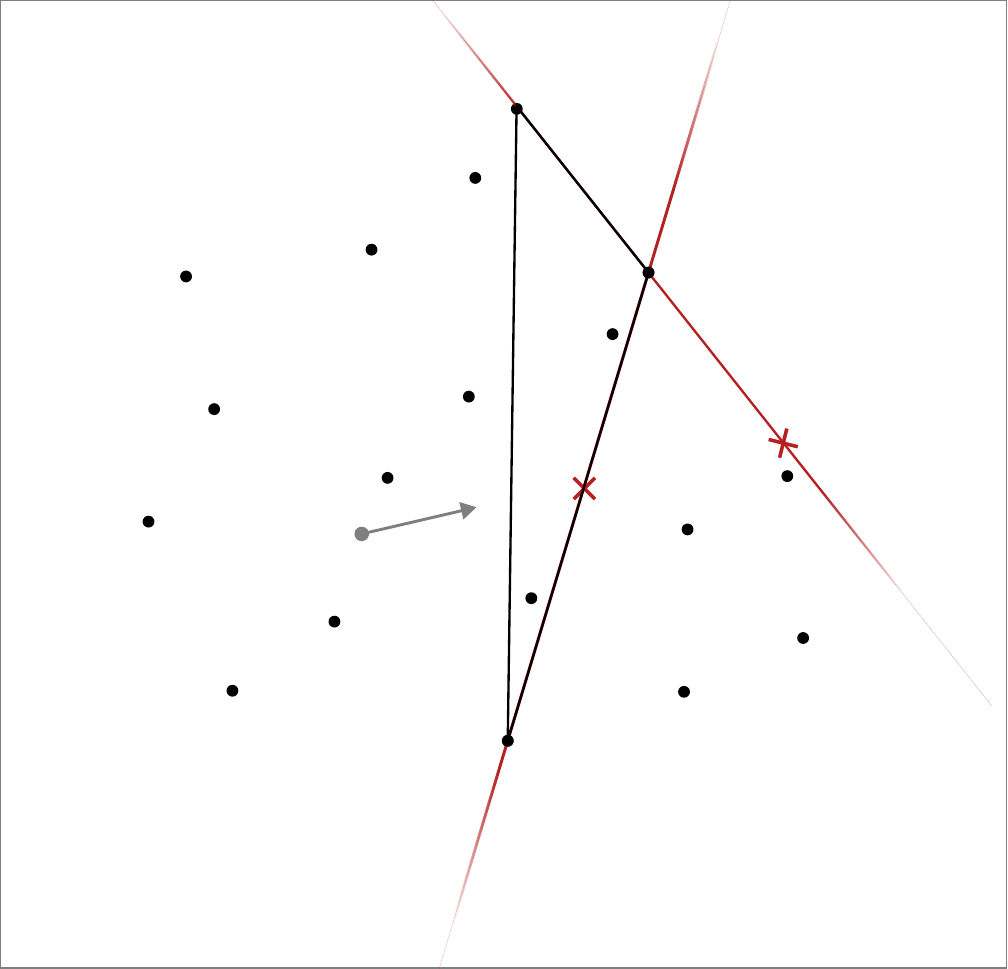
        \caption{Intersection with border hyperplanes in the right direction.}
        \label{subfig:extend_simplex_4}
      \end{subfigure}
      
      \begin{subfigure}{0.49\columnwidth}
        \def\svgwidth{\linewidth}
        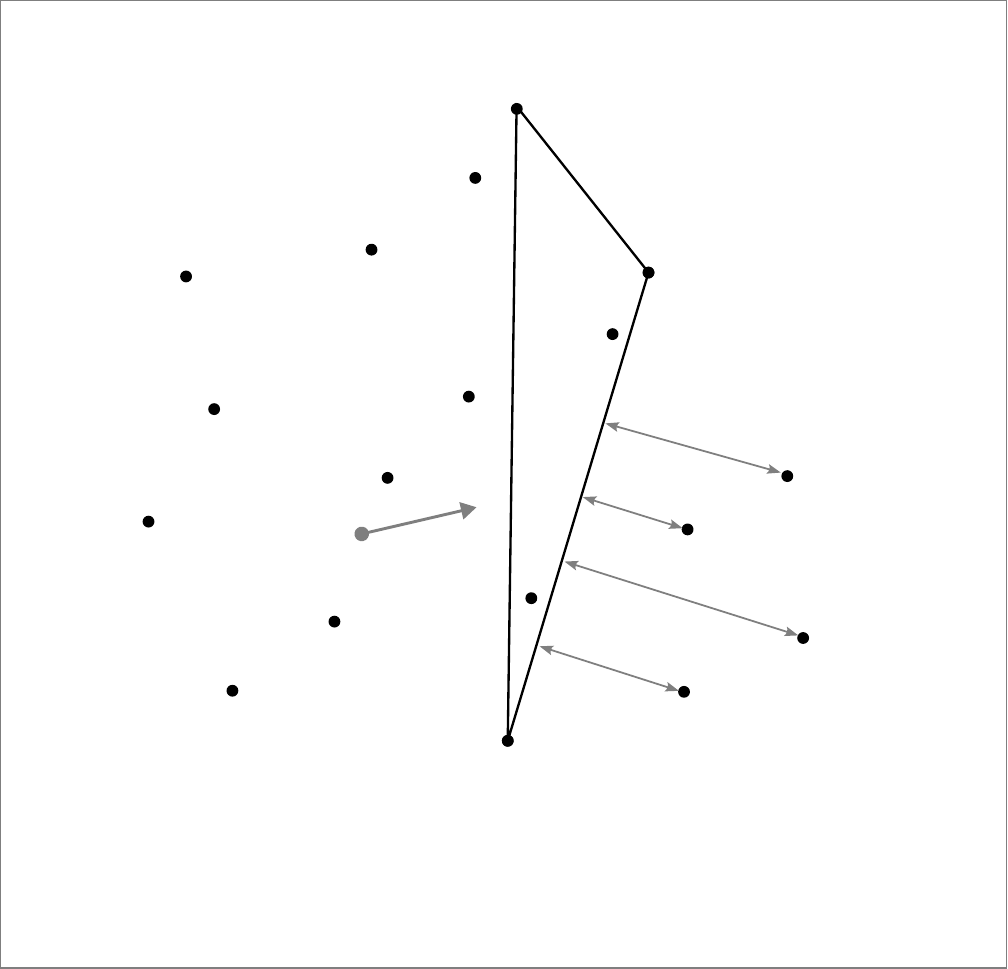
        \caption{Extreme point selection in facet normal direction.}
        \label{subfig:extend_simplex_5}
      \end{subfigure}
      \begin{subfigure}{0.49\columnwidth}
        \def\svgwidth{\linewidth}
        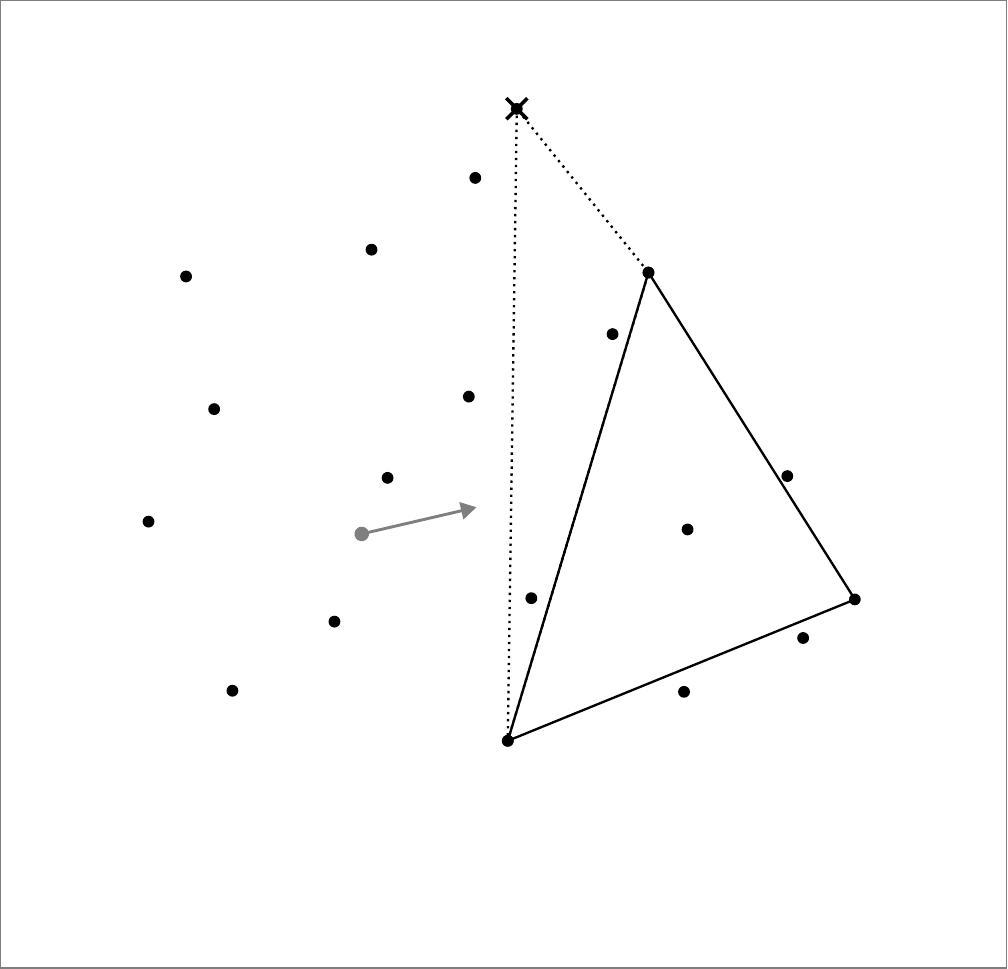
        \caption{Creation of an improved simplex.}
        \label{subfig:extend_simplex_6}
      \end{subfigure}

    \caption{Single iteration step for moving a simplex towards the bordering facet intersecting with the goal wrench.}
    \label{fig:simplex_extension}
\end{figure}

\subsection{Iterative Extension of the Simplex towards the Hull Intersection}
Given the initial simplex containing the origin, we need to extend it so that it contains the outer border of the Minkowski Volume and also intersects with $\mathbf{w}_{goal}$.
The algorithm works in the way that the facet of the current simplex intersecting with $\mathbf{w}_{goal}$ is extended along this facet's normal $\mathbf{n}$ towards the border.
One iteration loop of this algorithm is visualized in Fig.~\ref{fig:simplex_extension}.
For this, the normal vectors for each simplex facet are calculated (Fig.~\ref{subfig:extend_simplex_2}).
The normal vectors that do not point in the direction of the ray and thereby can only be entrance faces to the simplex are discarded (Fig.~\ref{subfig:extend_simplex_4}).
The hyperplane that has the closest ray intersection point to the origin is then extended.
The extension is done by again finding the extreme wrench $\mathbf{w}_{max}$ from the normal direction $\mathbf{n}$ of the exit facet (Fig.~\ref{subfig:extend_simplex_5}).
This extreme point replaces the wrench point that was not connected to the exit facet (Fig.~\ref{subfig:extend_simplex_6}).
The algorithm will terminate as soon as the exit facet is part of the boundary of the Minkowski volume $FW$.
This can be validated by checking if the new extreme point in the normal direction is already part of the facet.

\subsection{Calculation of Joint Torque Actuation}
After applying these algorithms, the hull facet of $FW$ intersecting with the direction of $\mathbf{w}_{goal}$ is found.

As the hull represents the maximum possible wrenches available, the intersection of the direction of $\mathbf{w}_{goal}$ needs to be scaled down to the actual length of $\mathbf{w}_{goal}$.
These scaling transformations can be directly applied to the individual foot wrenches as defined in Equation~(\ref{eq:fw_consists_of_cmw}).
With this, the wrench for each foot is known and can be transformed into the respective joint torques using the inverse dynamics transform $T_{ID}$.

It should be noted that all algorithms consist of only simple numerical calculations and matrix multiplications on the processor side.
Through this, an efficient and fast calculation of joint torques is possible for arbitrary contact configurations while staying inside the joint limitations.

\section{RESULTS AND EXPERIMENTS}
\label{chap:results_and_experiments}

We evaluated the performance of the algorithm by deploying it on a six-legged robot in simulation and on real hardware to validate our proposed concept.
For the evaluation in simulation, we used MATLAB Simulink \cite{noauthor_matlab_2023}.
The hardware tests were evaluated on LAURON~VI \cite{eichmann_lauron_2025}.
The reason that a six-legged robot was chosen for the evaluation is that more contact points allow for higher stability, as more friction forces can be exerted.

\begin{figure}
    \centering
    \vspace{4pt}
    \includegraphics[width=0.47\textwidth, trim=3cm 2cm 2.5cm 1cm,clip]{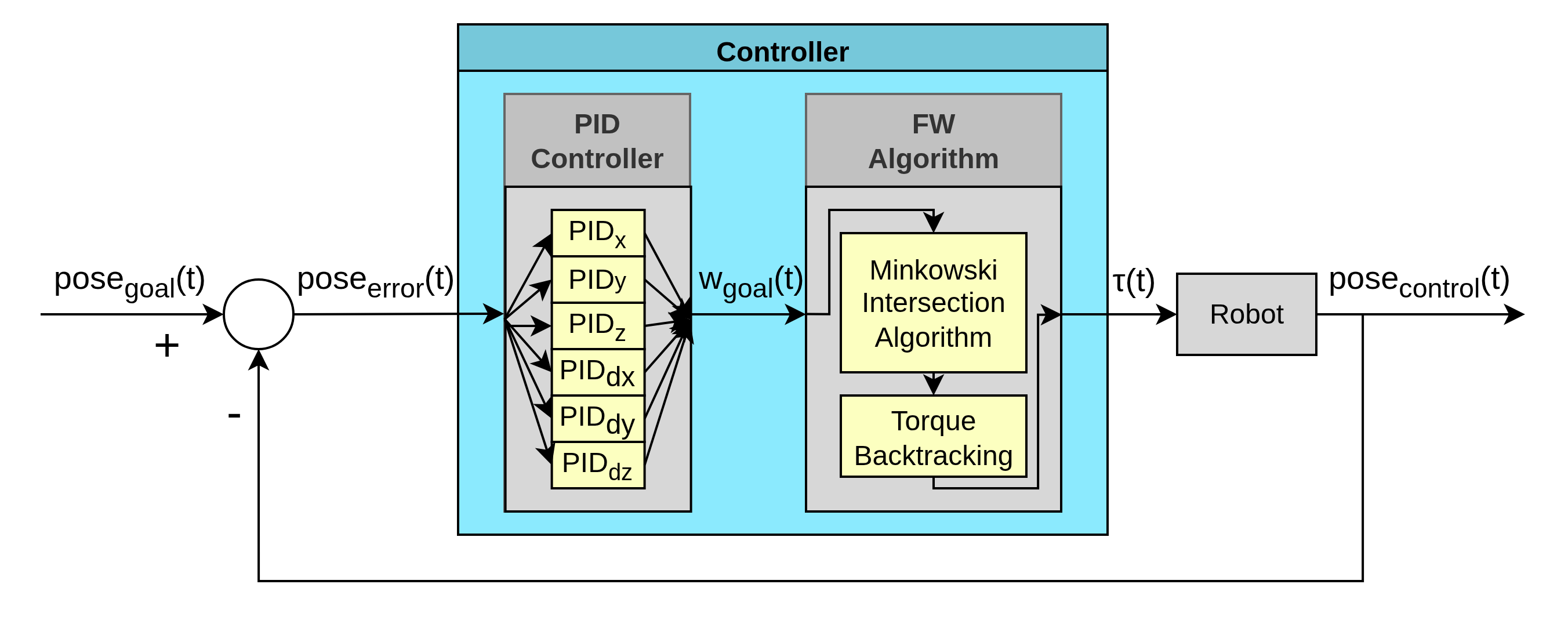}
    \caption{Controller architecture of the test runs.
 The error in each axis is transformed into a goal wrench for this axis.}
    \label{fig:architecture}
\end{figure}

As the algorithm can not run independently, an additional controller was implemented to track a pose with the robot's body by controlling wrenches on the base to follow the given pose.
The error was then forwarded into a set of PID controllers.
The basic architecture of the controller is shown in Fig.~\ref{fig:architecture}.
The whole control loop is running on a frequency of $50\ Hz$ in a real-time simulation.

To evaluate the algorithm, the ability to control body poses was tested in simulation with the results being presented in Section~\ref{chap:eval_stable_pose} and Section~\ref{chap:eval_actuation}.
The setup for each evaluation scenario is done by spawning the robot in the configuration that provides contact with all surfaces.
The friction model for the contacts corresponds to the Coulomb friction with a $FP_c$ consisting of $44$ vertices for each contact.

Besides the simulation, a hardware test was conducted to validate the simulation results, as shown in Section~\ref{chap:hardware_test}.
For the hardware test, a row of steep wall contacts was chosen for the ease of construction and testing.
The hardware setup is shown in Fig.~\ref{fig:hardware_setup}.

\begin{figure*}[t] 
  \centering
  \vspace{4pt}
  \makebox[\textwidth][s]{%
  \begin{subfigure}{0.33\textwidth}
    \includegraphics[width=\linewidth]{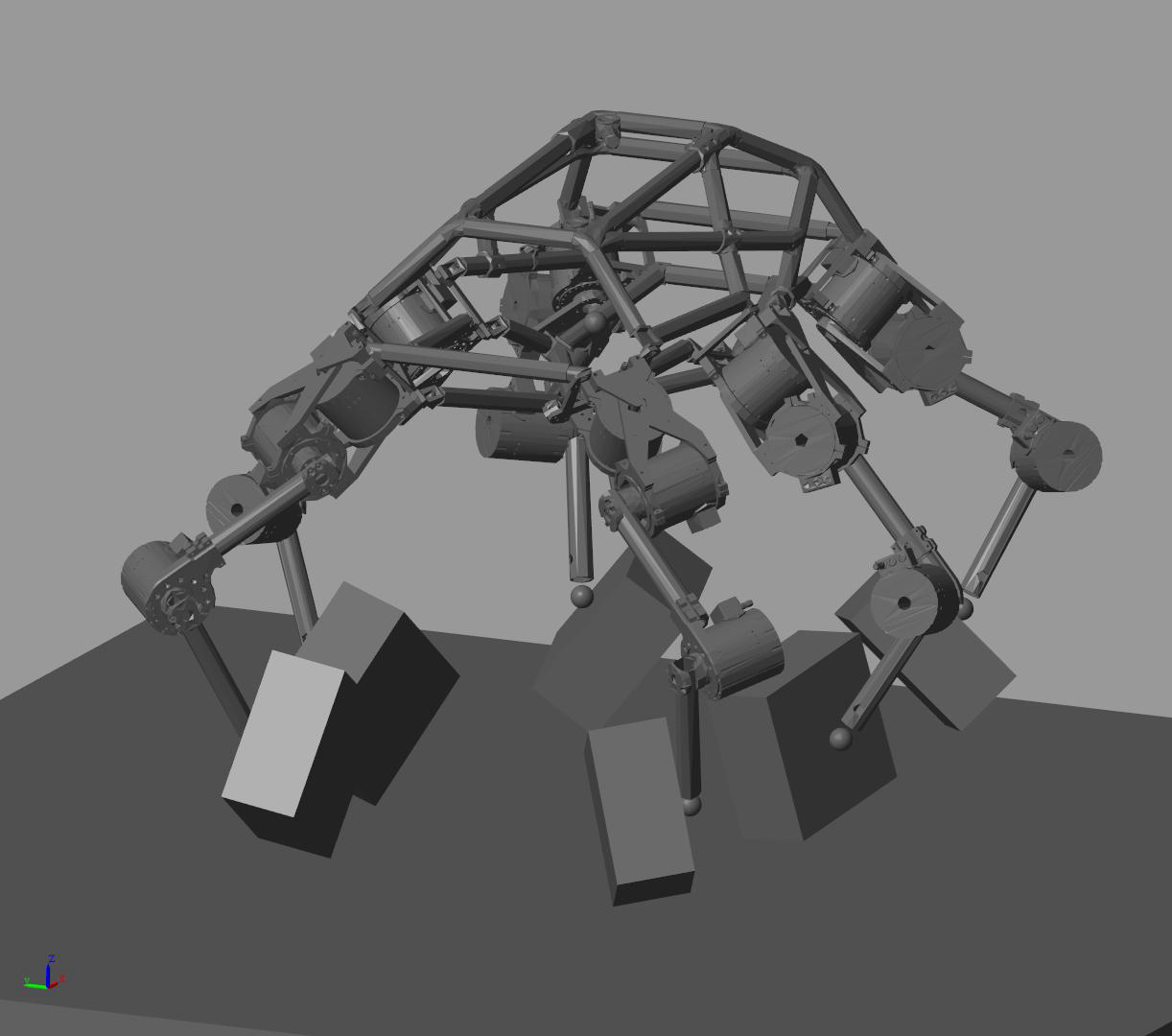}
  \end{subfigure}\hfill
  \begin{subfigure}{0.33\textwidth}
    \includegraphics[width=\linewidth]{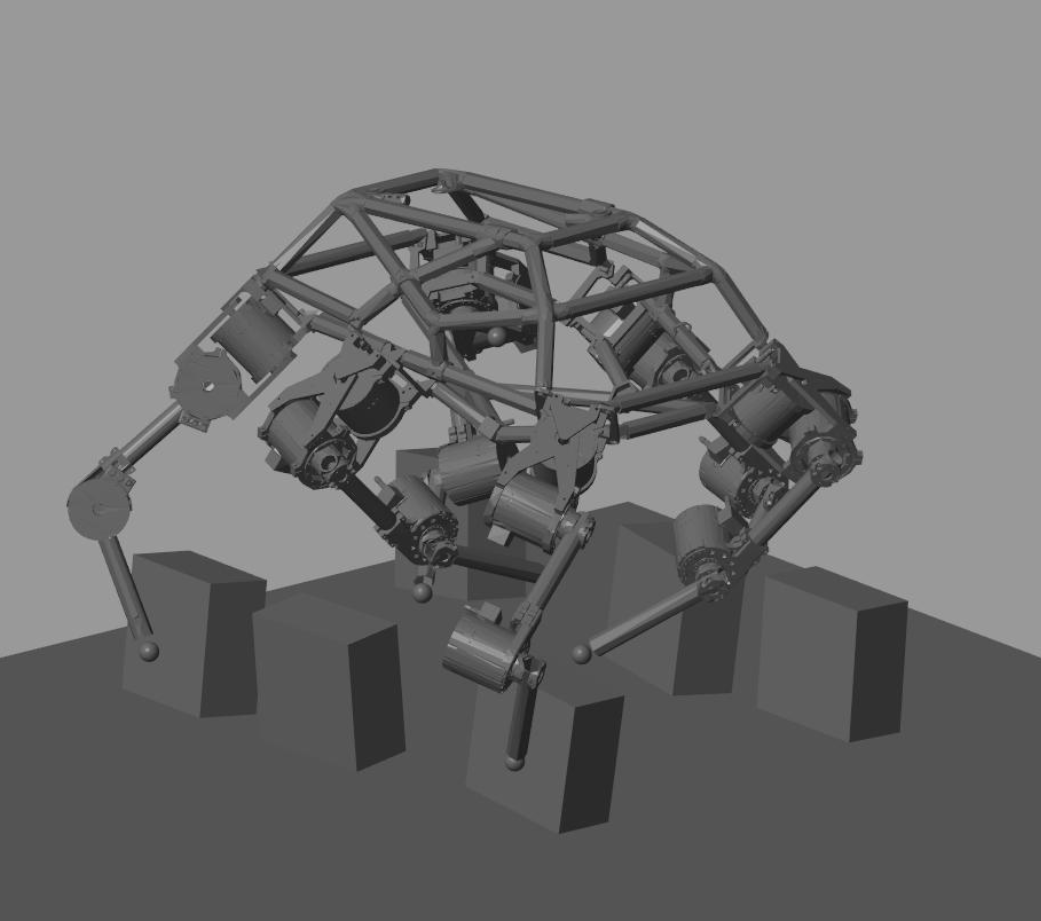}
  \end{subfigure}\hfill
  \begin{subfigure}{0.33\textwidth}
    \includegraphics[width=\linewidth]{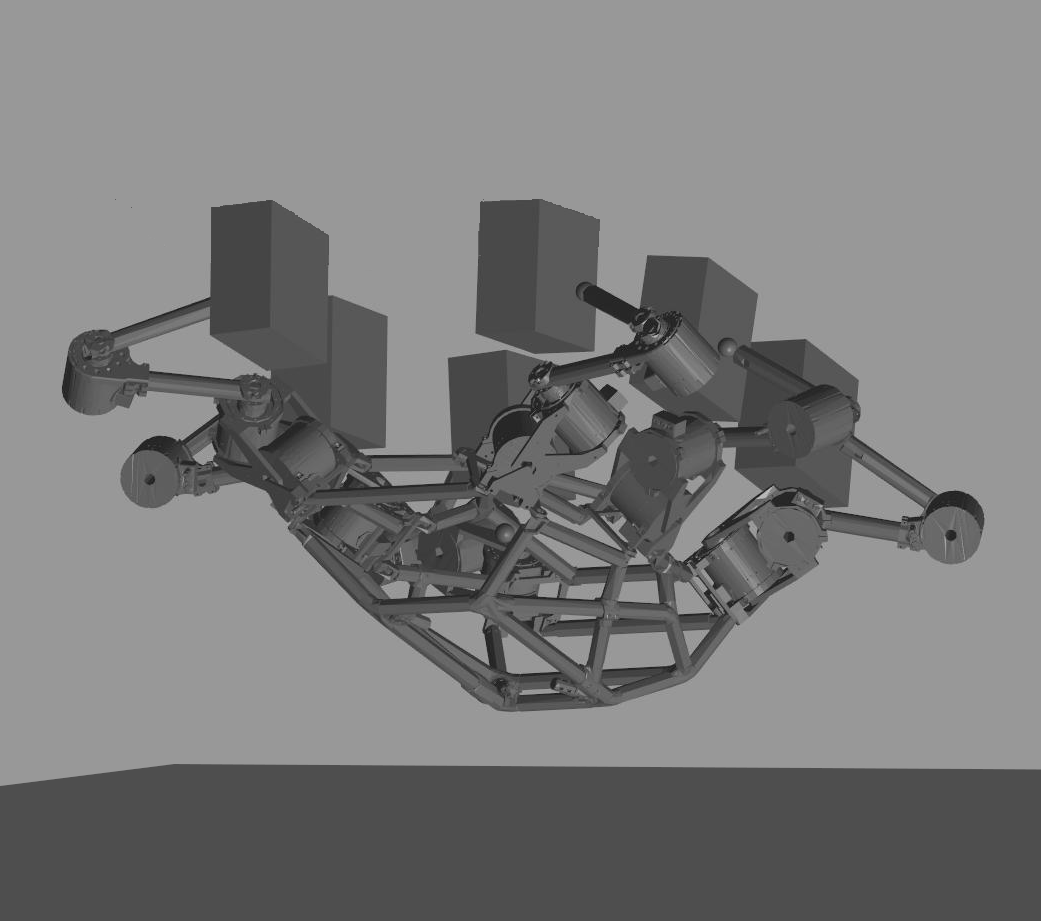}
  \end{subfigure}%
}

  \makebox[\textwidth][s]{%
  \begin{subfigure}{0.33\textwidth}
    \includegraphics[width=\linewidth]{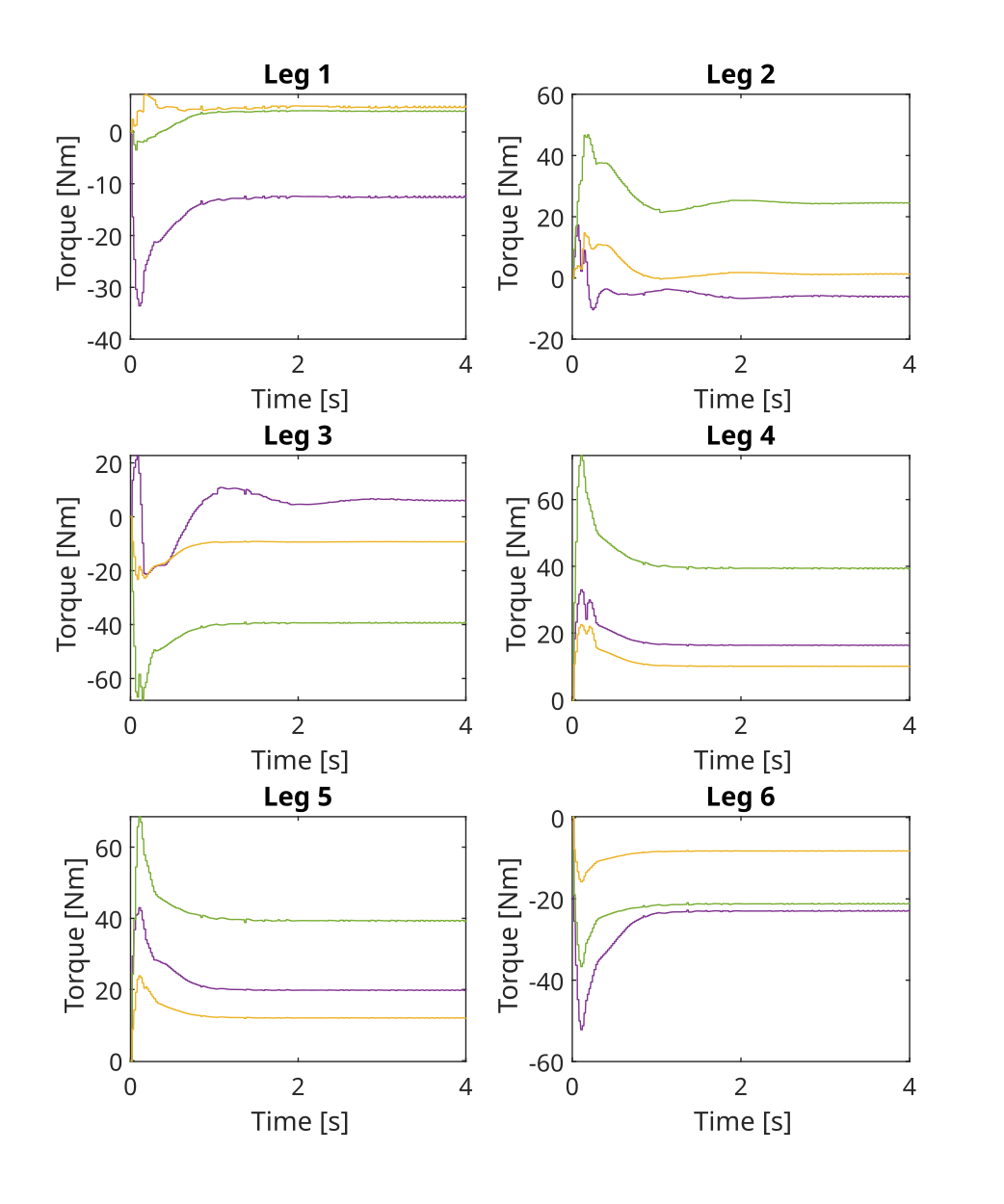}
    \caption{Randomized contact surfaces.}
  \end{subfigure}\hfill
  \begin{subfigure}{0.33\textwidth}
    \includegraphics[width=\linewidth]{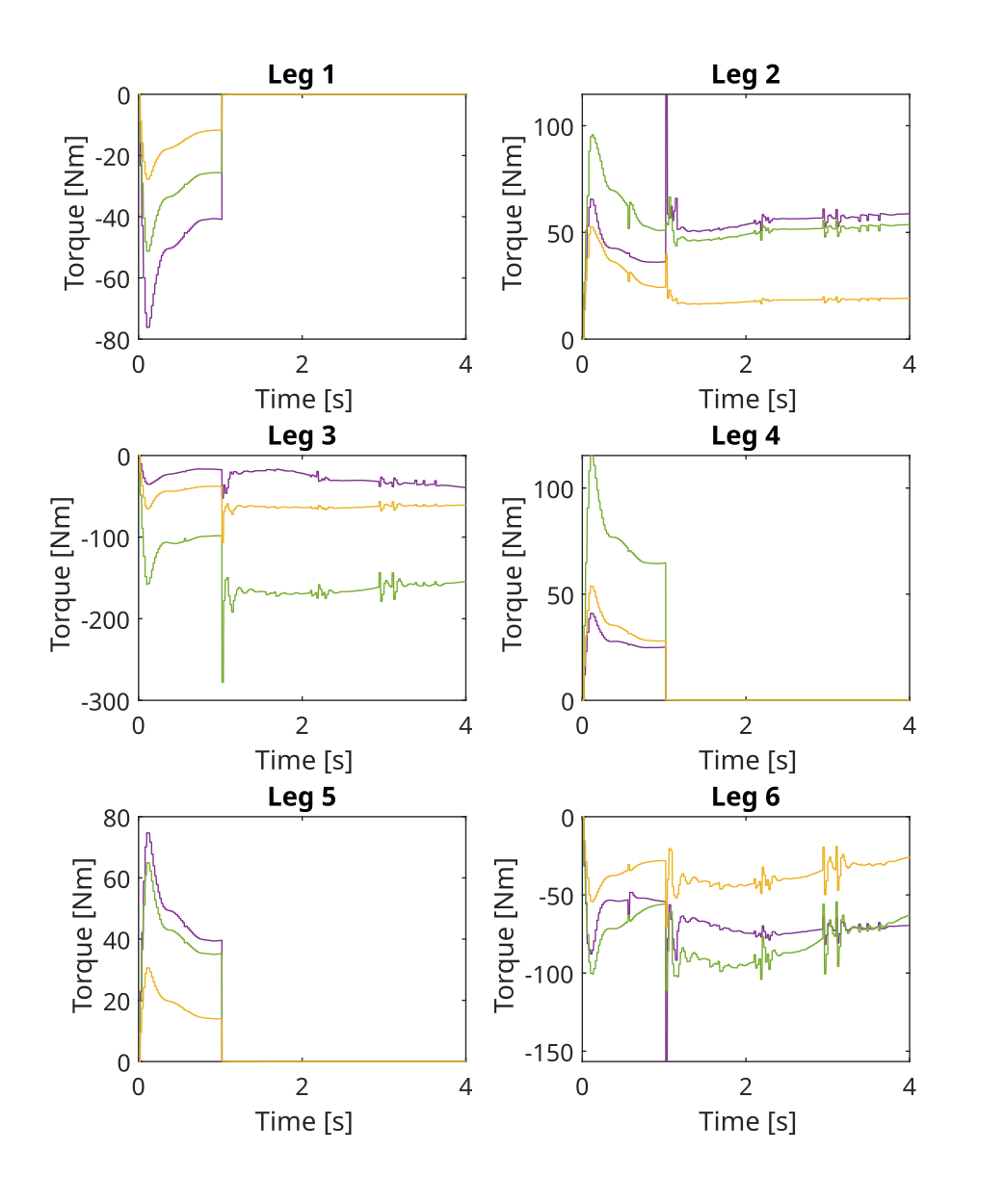}
    \caption{Loss of contact in three feet.}
    \label{fig:stable_overinclined}
  \end{subfigure}\hfill
  \begin{subfigure}{0.33\textwidth}
    \includegraphics[width=\linewidth]{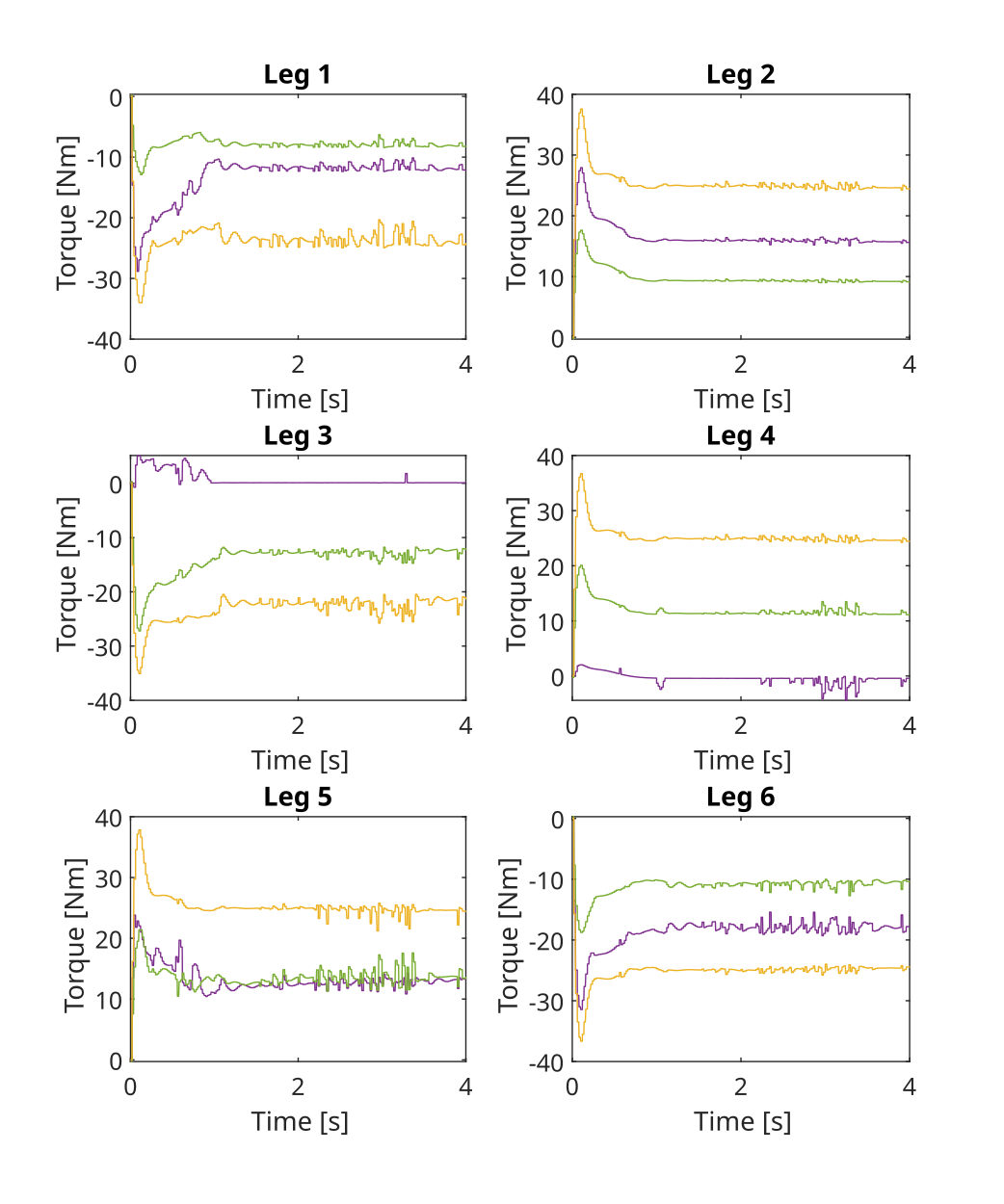}
    \caption{Overhead hang on vertical surfaces.}
  \end{subfigure}%
}
  
  \caption{Different simulated scenarios in which the robot was activated in the above shown position and had to keep itself stable.
 The applied torques for each leg are shown below, with the coloring indicating the three motors in each leg.}
  \label{fig:test_scenarios_stable}
\end{figure*}

\subsection{Stable Pose}
\label{chap:eval_stable_pose}
Multiple scenarios have been evaluated to test the algorithm's capabilities in different poses as listed in Table~\ref{tab:test_scenarios_short}.
Flat ground setups have also been tested, but to showcase the algorithm's capabilities, more complex scenarios were selected.
In Table~\ref{tab:test_scenarios_short}, ten different test setups are shown in three main test categories.
For each scenario, the time until a stable steady-state was reached is shown in the last column.
An interesting result is that the asymmetry in the loss of contact scenarios led to a bigger swinging motion that was harder to compensate than a loss of more contacts that was symmetrical. 
Three additional evaluation scenarios are shown in detail in Fig.~\ref{fig:test_scenarios_stable} with the torques applied by each leg and each motor, respectively.
Above each plot, the corresponding test setup of the robot for each simulated scenario is shown.
The torque plots show no major oscillations in the commanded joint torques, except for a slight jitter in extreme situations like the overhead hang and the contact loss of three legs.

These test results show that the proposed approach for calculating joint torques, as well as the pose controller, are well-suited for finding and applying contact forces in arbitrary environments to keep a stable robot pose.

\subsection{Body Pose Control}
\label{chap:eval_actuation}
\begin{figure}
  \centering
  \vspace{4pt}
  \makebox[0.5\textwidth][s]{%
  \begin{subfigure}{0.24\textwidth}
    \includegraphics[width=\linewidth]{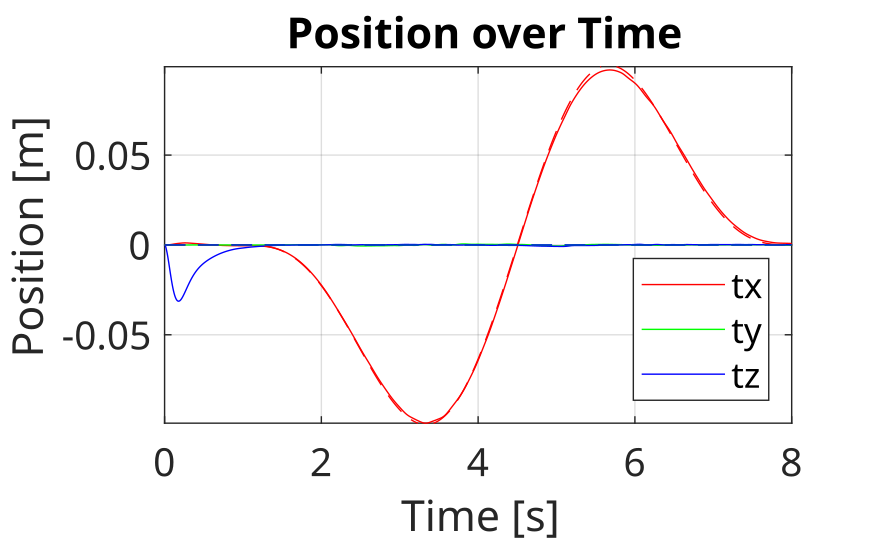}
    \caption{Movement along the x-axis.}
    \label{subfig:x_movement}
  \end{subfigure}
  \begin{subfigure}{0.24\textwidth}
    \includegraphics[width=\linewidth]{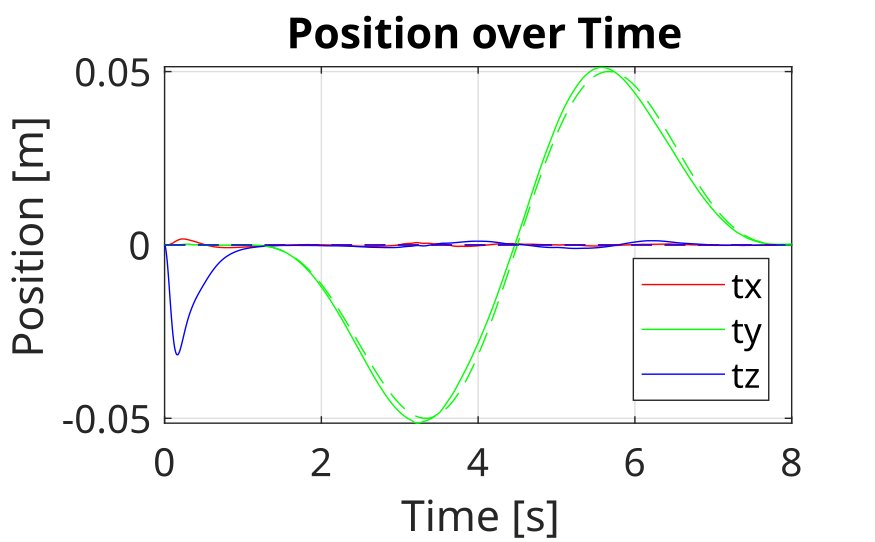}
    \caption{Movement along the y-axis.}
    \label{subfig:y_movement}
  \end{subfigure}
}
\makebox[0.5\textwidth][s]{%
  \begin{subfigure}{0.24\textwidth}
    \includegraphics[width=\linewidth]{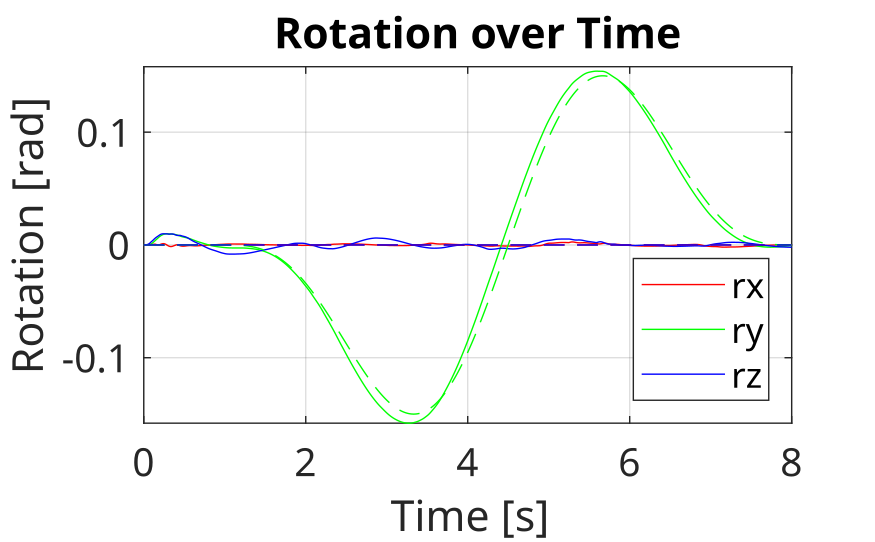}
    \caption{Rotation around the y-axis.}
    \label{subfig:y_rotation}
  \end{subfigure}%
  \begin{subfigure}{0.24\textwidth}
    \includegraphics[width=\linewidth]{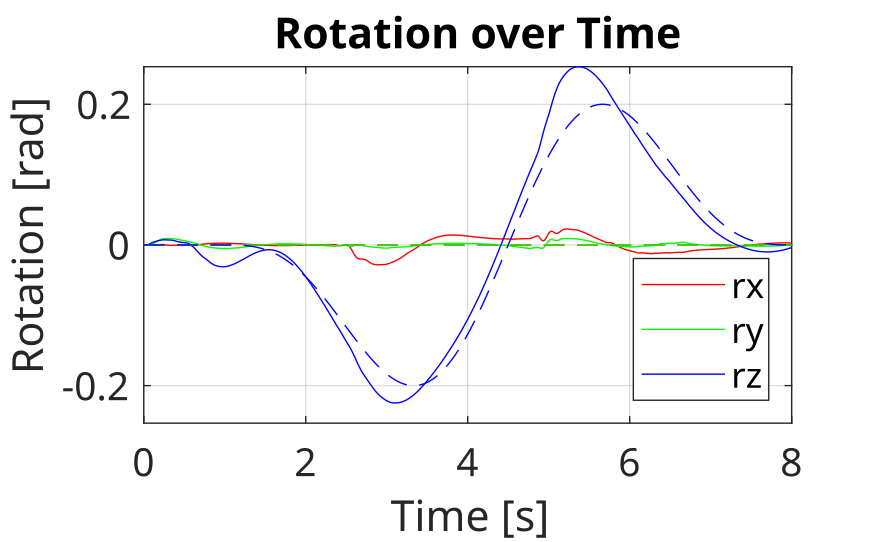}
    \caption{Rotation around the z-axis.}
    \label{subfig:z_rotation}
  \end{subfigure}%
}
\caption{Controlled position and rotation over time.
Each plot contains one individual test run of a pose control, with one plotted line for the change along each axis.
The position displacements in (a) and (b) show the actual position of the robot base with regard to its default pose.
The rotation plots (c) and (d) do the same, but for the rotational axes.
The dashed line in each plot shows the goal value for each axis.}
\label{fig:actuation_test}
\end{figure}

Besides keeping a stable pose, the algorithm also allows for calculating contact forces to actuate a specific wrench on the robot's base.
Through the controller shown in Fig.~\ref{fig:architecture}, a new goal pose for the robot can be defined.
To evaluate the actuation capabilities, movements were commanded on all axes to be followed by the robot.
The test setup consists of purely vertical contact surfaces.
In Fig.~\ref{fig:actuation_test}, the results of the displacement test are shown.
After stabilizing the robot's pose after the spawn drop, a sinusoidal movement for the robot base was commanded.  
As shown in the plots Fig.~\ref{subfig:x_movement} and Fig.~\ref{subfig:y_movement}, movements in the x- and y-axis directions are actuated with a small margin, closely following the goal position.
This movement corresponds to a shifting of the robot's center of mass within its stable boundary.
The rotation plots Fig.~\ref{subfig:y_rotation} and Fig.~\ref{subfig:z_rotation} show the rotation of the robot in accordance with the given goal pose.
Here, the controller followed the trajectory along the z-axis rotation with a higher offset.
The reason is that a rotation in the z-axis moves the robot base closer to the boundaries of the actuation limits in the Minkowski sum.
So instead of applying high forces to follow the goal curve exactly, the controller prioritized staying within stable limits instead. 

The results show that the controller is able to follow the given actuation command closely.
Even border cases like the z-axis rotation can be stably actuated, without losing control of the robot.
That means that within the boundaries of the $FW$, the position and orientation can be freely controlled.
This way, the robot can also shift the $CoM$, which is necessary to enable a stable stepping motion.

\begin{table}
  \centering
  \vspace{4pt}
  \caption{A selection of test setups used for the evaluation of the algorithm.
 For each test, the time until the robot reached a steady state is shown.}
  \label{tab:test_scenarios_short}
  \begin{tabular}{|c|l|c|}
    \hline
    & Scenario & Time until stable \\
    \hline
    \multirow{4}{*}{\makecell{Contact Surface\\Setup}}
    &Randomized Contacts & $1.7s$ \\
    &Vertical Contacts & $0.3s$ \\
    &Overinclined Contacts & $0.8s$ \\
    &Overhead Contacts & $0.9s$ \\
    \hline
    \multirow{4}{*}{\makecell{Simulated Control\\Frequency}}
    &Controller at $20\ Hz$ & $-$ \\
    &Controller at $35\ Hz$ & $0.6s$ \\
    &Controller at $50\ Hz$ & $0.7s$ \\
    &Controller at $200\ Hz$ & $0.7s$ \\
    \hline
    \multirow{2}{*}{\makecell{Contact Loss of Legs}}
    &Loss of 1 contact &  $2.0s$ \\
    &Loss of 3 contacts & $0.2s$ \\
    \hline
  \end{tabular}
\end{table}

\subsection{Controller evaluation on robotic hardware}
\label{chap:hardware_test}
The algorithm for wrench calculations was also tested on the real robotic hardware.
For this, the six-legged robot LAURON~VI \cite{eichmann_lauron_2025} was used.
For these evaluations, the software stack was migrated from the MATLAB Simulink Simulation \cite{noauthor_matlab_2023} to the actual robot.
The robot uses the ROS 2 Framework \cite{macenski_robot_2022} for communication purposes, while the implementation of our algorithm was done using the Pinocchio C++ framework for robot dynamics \cite{carpentier_pinocchio_2019} and the quickhull algorithm \cite{barber_quickhull_1996}.
An additional position controller was used as part of the control loop on the motors.
This was necessary, as during the test, the proposed torque controller only relies on friction to keep the contact points, but does not hold the position in case of slippage.
The algorithm was run on an Intel(R) Core(TM) i7-11800H 2.30GHz CPU with 32GB of RAM.
As this setup has less calculation power than the simulation setup, the execution frequency dropped from the previously evaluated $50\ Hz$ to $45\ Hz$-$49\ Hz$ during the tests on the real robot.
The reason is mostly due to the not fully optimized migration from MATLAB to C++.
The controller was tested on a steep wall scenario as shown in Fig.~\ref{fig:hardware_setup}.
The evaluation was done by setting the robot onto the wall using a crane, and starting our controller while having the feet in contact with the wall.

This scenario was tested with different PID controller gains in a qualitative analysis.
Each run was tested with different PID configurations, varying high and low P-Gains as well as D-Settings.
The position controller's P-Gain was then lowered until slippage was imminent.
While every test run was able to keep the robot stable in its setup configuration on the wall, higher P-Gains create the tendency for high-frequency oscillation in the joint controllers.
This results in the robot losing grip when exerting more force on a contact point to move the robot's body pose.
Lower values for the P-Gain, however, create too little force to move the robot's body and instead slide along the contact surface downwards.
Additional disturbance tests were also conducted by pushing the robot while in the initial configuration.
Even strong pushes that rotated the robot's base were able to be counteracted while keeping contact with the wall.
Videos of the hardware tests are available under (the link will be added for the final submission after acceptance confirmation).

Besides the stability evaluation, we evaluated the contribution of our proposed controller against the additionally used position controller.
To do that, the joint torques for each joint were compared to the commanded values of our controller, and the median of the percentage shares was analyzed.
This resulted in our controller publishing a percentage of the final torque of $70\%$-$80\%$.
This shows that even with the supporting position controller, the main torque commands are still set through our proposed stability controller.

The proof-of-concept tests showed that the controller created correct wrench commands for each contact point and kept the robot stable even in very difficult contact scenarios.
With a frequency of $47\ Hz$, which corresponds to a calculation time of $21.2\ ms$, it achieves a significant speedup in comparison to a similarly detailed baseline \cite{orsolino_application_2018}, which takes $350\ ms$ for the stability calculation of a simpler four-contact scenario.

\section{CONCLUSIONS AND FUTURE WORKS}
\label{chap:conclusion_and_future_works}
This paper proposed a novel algorithm to efficiently calculate a robot's body and foot contact wrenches.
The shown algorithmic improvements in calculating only a small fraction of the full Minkowski sum were implemented and validated successfully on a real robot. 
The algorithm was able to evaluate a full friction model together with joint actuation limits to generate feasible contact forces within a $50\ Hz$ frequency.
A controller was proposed using this calculation algorithm for real-time control of the six-legged robot LAURON~VI, achieving a control frequency of $45$-$49\ Hz$.
The controller was tested extensively in simulation for a multitude of different contact configurations, as well as evaluated on the real robot.
As shown in the evaluation section, the controller was able to keep the real robot stable in a challenging environment.

Future work on this controller includes improving the performance of the controller and the algorithm by using more parallelization.
The stability criterion will be further incorporated into reinforcement learning approaches to train climbing behavior and improve stable movement.

\bibliographystyle{IEEEtran}
\bibliography{IEEEabrv,library_new.bib}

\end{document}

%% file: 1.pdf_tex
\begingroup%
  \makeatletter%
  \providecommand\color[2][]{%
    \errmessage{(Inkscape) Color is used for the text in Inkscape, but the package 'color.sty' is not loaded}%
    \renewcommand\color[2][]{}%
  }%
  \providecommand\transparent[1]{%
    \errmessage{(Inkscape) Transparency is used (non-zero) for the text in Inkscape, but the package 'transparent.sty' is not loaded}%
    \renewcommand\transparent[1]{}%
  }%
  \providecommand\rotatebox[2]{#2}%
  \newcommand*\fsize{\dimexpr\f@size pt\relax}%
  \newcommand*\lineheight[1]{\fontsize{\fsize}{#1\fsize}\selectfont}%
  \ifx\svgwidth\undefined%
    \setlength{\unitlength}{483.57504056bp}%
    \ifx\svgscale\undefined%
      \relax%
    \else%
      \setlength{\unitlength}{\unitlength * \real{\svgscale}}%
    \fi%
  \else%
    \setlength{\unitlength}{\svgwidth}%
  \fi%
  \global\let\svgwidth\undefined%
  \global\let\svgscale\undefined%
  \makeatother%
  \begin{picture}(1,0.96141066)%
    \lineheight{1}%
    \setlength\tabcolsep{0pt}%
    \put(0,0){\includegraphics[width=\unitlength,page=1]{1.pdf}}%
    \put(0.22520601,0.17846342){\color[rgb]{0.71372549,0.12941176,0.12941176}\makebox(0,0)[lt]{\lineheight{1.25}\smash{\begin{tabular}[t]{l}$origin$\end{tabular}}}}%
    \put(0,0){\includegraphics[width=\unitlength,page=2]{1.pdf}}%
  \end{picture}%
\endgroup%

%% file: 2.pdf_tex
\begingroup%
  \makeatletter%
  \providecommand\color[2][]{%
    \errmessage{(Inkscape) Color is used for the text in Inkscape, but the package 'color.sty' is not loaded}%
    \renewcommand\color[2][]{}%
  }%
  \providecommand\transparent[1]{%
    \errmessage{(Inkscape) Transparency is used (non-zero) for the text in Inkscape, but the package 'transparent.sty' is not loaded}%
    \renewcommand\transparent[1]{}%
  }%
  \providecommand\rotatebox[2]{#2}%
  \newcommand*\fsize{\dimexpr\f@size pt\relax}%
  \newcommand*\lineheight[1]{\fontsize{\fsize}{#1\fsize}\selectfont}%
  \ifx\svgwidth\undefined%
    \setlength{\unitlength}{483.57504056bp}%
    \ifx\svgscale\undefined%
      \relax%
    \else%
      \setlength{\unitlength}{\unitlength * \real{\svgscale}}%
    \fi%
  \else%
    \setlength{\unitlength}{\svgwidth}%
  \fi%
  \global\let\svgwidth\undefined%
  \global\let\svgscale\undefined%
  \makeatother%
  \begin{picture}(1,0.96141066)%
    \lineheight{1}%
    \setlength\tabcolsep{0pt}%
    \put(0,0){\includegraphics[width=\unitlength,page=1]{2.pdf}}%
    \put(0.22520601,0.17846342){\color[rgb]{0.71372549,0.12941176,0.12941176}\makebox(0,0)[lt]{\lineheight{1.25}\smash{\begin{tabular}[t]{l}$origin$\end{tabular}}}}%
    \put(0,0){\includegraphics[width=\unitlength,page=2]{2.pdf}}%
    \put(0.5688943,0.80055119){\color[rgb]{0.71372549,0.1254902,0.1254902}\makebox(0,0)[lt]{\lineheight{1.25}\smash{\begin{tabular}[t]{l}$\mathbf{n}_1$\end{tabular}}}}%
    \put(0.61862913,0.49317338){\color[rgb]{0.70588235,0.13333333,0.13333333}\makebox(0,0)[lt]{\lineheight{1.25}\smash{\begin{tabular}[t]{l}$\mathbf{n}_2$\end{tabular}}}}%
    \put(0,0){\includegraphics[width=\unitlength,page=3]{2.pdf}}%
    \put(0.33632847,0.61550201){\color[rgb]{0.71372549,0.12941176,0.12941176}\makebox(0,0)[lt]{\lineheight{1.25}\smash{\begin{tabular}[t]{l}$\mathbf{n}_3$\end{tabular}}}}%
    \put(0,0){\includegraphics[width=\unitlength,page=4]{2.pdf}}%
  \end{picture}%
\endgroup%

%% file: 3.pdf_tex
\begingroup%
  \makeatletter%
  \providecommand\color[2][]{%
    \errmessage{(Inkscape) Color is used for the text in Inkscape, but the package 'color.sty' is not loaded}%
    \renewcommand\color[2][]{}%
  }%
  \providecommand\transparent[1]{%
    \errmessage{(Inkscape) Transparency is used (non-zero) for the text in Inkscape, but the package 'transparent.sty' is not loaded}%
    \renewcommand\transparent[1]{}%
  }%
  \providecommand\rotatebox[2]{#2}%
  \newcommand*\fsize{\dimexpr\f@size pt\relax}%
  \newcommand*\lineheight[1]{\fontsize{\fsize}{#1\fsize}\selectfont}%
  \ifx\svgwidth\undefined%
    \setlength{\unitlength}{483.57504056bp}%
    \ifx\svgscale\undefined%
      \relax%
    \else%
      \setlength{\unitlength}{\unitlength * \real{\svgscale}}%
    \fi%
  \else%
    \setlength{\unitlength}{\svgwidth}%
  \fi%
  \global\let\svgwidth\undefined%
  \global\let\svgscale\undefined%
  \makeatother%
  \begin{picture}(1,0.96141066)%
    \lineheight{1}%
    \setlength\tabcolsep{0pt}%
    \put(0,0){\includegraphics[width=\unitlength,page=1]{3.pdf}}%
    \put(0.22520601,0.17846342){\color[rgb]{0.71372549,0.12941176,0.12941176}\makebox(0,0)[lt]{\lineheight{1.25}\smash{\begin{tabular}[t]{l}$origin$\end{tabular}}}}%
    \put(0,0){\includegraphics[width=\unitlength,page=2]{3.pdf}}%
    \put(0.33632847,0.61550201){\color[rgb]{0.71372549,0.12941176,0.12941176}\makebox(0,0)[lt]{\lineheight{1.25}\smash{\begin{tabular}[t]{l}$\mathbf{n}_3$\end{tabular}}}}%
    \put(0,0){\includegraphics[width=\unitlength,page=3]{3.pdf}}%
  \end{picture}%
\endgroup%

%% file: 4.pdf_tex
\begingroup%
  \makeatletter%
  \providecommand\color[2][]{%
    \errmessage{(Inkscape) Color is used for the text in Inkscape, but the package 'color.sty' is not loaded}%
    \renewcommand\color[2][]{}%
  }%
  \providecommand\transparent[1]{%
    \errmessage{(Inkscape) Transparency is used (non-zero) for the text in Inkscape, but the package 'transparent.sty' is not loaded}%
    \renewcommand\transparent[1]{}%
  }%
  \providecommand\rotatebox[2]{#2}%
  \newcommand*\fsize{\dimexpr\f@size pt\relax}%
  \newcommand*\lineheight[1]{\fontsize{\fsize}{#1\fsize}\selectfont}%
  \ifx\svgwidth\undefined%
    \setlength{\unitlength}{483.57504056bp}%
    \ifx\svgscale\undefined%
      \relax%
    \else%
      \setlength{\unitlength}{\unitlength * \real{\svgscale}}%
    \fi%
  \else%
    \setlength{\unitlength}{\svgwidth}%
  \fi%
  \global\let\svgwidth\undefined%
  \global\let\svgscale\undefined%
  \makeatother%
  \begin{picture}(1,0.96141066)%
    \lineheight{1}%
    \setlength\tabcolsep{0pt}%
    \put(0,0){\includegraphics[width=\unitlength,page=1]{4.pdf}}%
    \put(0.22520601,0.17846342){\color[rgb]{0.71372549,0.12941176,0.12941176}\makebox(0,0)[lt]{\lineheight{1.25}\smash{\begin{tabular}[t]{l}$origin$\end{tabular}}}}%
    \put(0,0){\includegraphics[width=\unitlength,page=2]{4.pdf}}%
  \end{picture}%
\endgroup%

%% file: origin_dir.pdf_tex
\begingroup%
  \makeatletter%
  \providecommand\color[2][]{%
    \errmessage{(Inkscape) Color is used for the text in Inkscape, but the package 'color.sty' is not loaded}%
    \renewcommand\color[2][]{}%
  }%
  \providecommand\transparent[1]{%
    \errmessage{(Inkscape) Transparency is used (non-zero) for the text in Inkscape, but the package 'transparent.sty' is not loaded}%
    \renewcommand\transparent[1]{}%
  }%
  \providecommand\rotatebox[2]{#2}%
  \newcommand*\fsize{\dimexpr\f@size pt\relax}%
  \newcommand*\lineheight[1]{\fontsize{\fsize}{#1\fsize}\selectfont}%
  \ifx\svgwidth\undefined%
    \setlength{\unitlength}{483.57504056bp}%
    \ifx\svgscale\undefined%
      \relax%
    \else%
      \setlength{\unitlength}{\unitlength * \real{\svgscale}}%
    \fi%
  \else%
    \setlength{\unitlength}{\svgwidth}%
  \fi%
  \global\let\svgwidth\undefined%
  \global\let\svgscale\undefined%
  \makeatother%
  \begin{picture}(1,0.96141066)%
    \lineheight{1}%
    \setlength\tabcolsep{0pt}%
    \put(0,0){\includegraphics[width=\unitlength,page=1]{origin_dir.pdf}}%
    \put(0.26539783,0.36737609){\color[rgb]{0.70980392,0.12941176,0.12941176}\makebox(0,0)[lt]{\lineheight{1.25}\smash{\begin{tabular}[t]{l}$origin$\end{tabular}}}}%
    \put(0.41601208,0.48290695){\color[rgb]{0.71372549,0.1254902,0.1254902}\makebox(0,0)[lt]{\lineheight{1.25}\smash{\begin{tabular}[t]{l}$\mathbf{w}_{goal}$\end{tabular}}}}%
  \end{picture}%
\endgroup%

%% file: simplex_clear.pdf_tex
\begingroup%
  \makeatletter%
  \providecommand\color[2][]{%
    \errmessage{(Inkscape) Color is used for the text in Inkscape, but the package 'color.sty' is not loaded}%
    \renewcommand\color[2][]{}%
  }%
  \providecommand\transparent[1]{%
    \errmessage{(Inkscape) Transparency is used (non-zero) for the text in Inkscape, but the package 'transparent.sty' is not loaded}%
    \renewcommand\transparent[1]{}%
  }%
  \providecommand\rotatebox[2]{#2}%
  \newcommand*\fsize{\dimexpr\f@size pt\relax}%
  \newcommand*\lineheight[1]{\fontsize{\fsize}{#1\fsize}\selectfont}%
  \ifx\svgwidth\undefined%
    \setlength{\unitlength}{483.57504056bp}%
    \ifx\svgscale\undefined%
      \relax%
    \else%
      \setlength{\unitlength}{\unitlength * \real{\svgscale}}%
    \fi%
  \else%
    \setlength{\unitlength}{\svgwidth}%
  \fi%
  \global\let\svgwidth\undefined%
  \global\let\svgscale\undefined%
  \makeatother%
  \begin{picture}(1,0.96141066)%
    \lineheight{1}%
    \setlength\tabcolsep{0pt}%
    \put(0,0){\includegraphics[width=\unitlength,page=1]{simplex_clear.pdf}}%
    \put(0.32492495,0.61891331){\color[rgb]{0.71372549,0.12941176,0.12941176}\makebox(0,0)[lt]{\lineheight{1.25}\smash{\begin{tabular}[t]{l}$\mathbf{n}_{1}$\end{tabular}}}}%
    \put(0.62346938,0.78698705){\color[rgb]{0.71372549,0.1254902,0.1254902}\makebox(0,0)[lt]{\lineheight{1.25}\smash{\begin{tabular}[t]{l}$\mathbf{n}_{2}$\end{tabular}}}}%
    \put(0.61692344,0.5098035){\color[rgb]{0.70588235,0.13333333,0.13333333}\makebox(0,0)[lt]{\lineheight{1.25}\smash{\begin{tabular}[t]{l}$\mathbf{n}_{3}$\end{tabular}}}}%
  \end{picture}%
\endgroup%

%% file: cutPoints1.pdf_tex
\begingroup%
  \makeatletter%
  \providecommand\color[2][]{%
    \errmessage{(Inkscape) Color is used for the text in Inkscape, but the package 'color.sty' is not loaded}%
    \renewcommand\color[2][]{}%
  }%
  \providecommand\transparent[1]{%
    \errmessage{(Inkscape) Transparency is used (non-zero) for the text in Inkscape, but the package 'transparent.sty' is not loaded}%
    \renewcommand\transparent[1]{}%
  }%
  \providecommand\rotatebox[2]{#2}%
  \newcommand*\fsize{\dimexpr\f@size pt\relax}%
  \newcommand*\lineheight[1]{\fontsize{\fsize}{#1\fsize}\selectfont}%
  \ifx\svgwidth\undefined%
    \setlength{\unitlength}{483.57504056bp}%
    \ifx\svgscale\undefined%
      \relax%
    \else%
      \setlength{\unitlength}{\unitlength * \real{\svgscale}}%
    \fi%
  \else%
    \setlength{\unitlength}{\svgwidth}%
  \fi%
  \global\let\svgwidth\undefined%
  \global\let\svgscale\undefined%
  \makeatother%
  \begin{picture}(1,0.96141066)%
    \lineheight{1}%
    \setlength\tabcolsep{0pt}%
    \put(0,0){\includegraphics[width=\unitlength,page=1]{cutPoints1.pdf}}%
  \end{picture}%
\endgroup%

%% file: cutPoints2.pdf_tex
\begingroup%
  \makeatletter%
  \providecommand\color[2][]{%
    \errmessage{(Inkscape) Color is used for the text in Inkscape, but the package 'color.sty' is not loaded}%
    \renewcommand\color[2][]{}%
  }%
  \providecommand\transparent[1]{%
    \errmessage{(Inkscape) Transparency is used (non-zero) for the text in Inkscape, but the package 'transparent.sty' is not loaded}%
    \renewcommand\transparent[1]{}%
  }%
  \providecommand\rotatebox[2]{#2}%
  \newcommand*\fsize{\dimexpr\f@size pt\relax}%
  \newcommand*\lineheight[1]{\fontsize{\fsize}{#1\fsize}\selectfont}%
  \ifx\svgwidth\undefined%
    \setlength{\unitlength}{483.57504056bp}%
    \ifx\svgscale\undefined%
      \relax%
    \else%
      \setlength{\unitlength}{\unitlength * \real{\svgscale}}%
    \fi%
  \else%
    \setlength{\unitlength}{\svgwidth}%
  \fi%
  \global\let\svgwidth\undefined%
  \global\let\svgscale\undefined%
  \makeatother%
  \begin{picture}(1,0.96141066)%
    \lineheight{1}%
    \setlength\tabcolsep{0pt}%
    \put(0,0){\includegraphics[width=\unitlength,page=1]{cutPoints2.pdf}}%
  \end{picture}%
\endgroup%

%% file: searchPoint.pdf_tex
\begingroup%
  \makeatletter%
  \providecommand\color[2][]{%
    \errmessage{(Inkscape) Color is used for the text in Inkscape, but the package 'color.sty' is not loaded}%
    \renewcommand\color[2][]{}%
  }%
  \providecommand\transparent[1]{%
    \errmessage{(Inkscape) Transparency is used (non-zero) for the text in Inkscape, but the package 'transparent.sty' is not loaded}%
    \renewcommand\transparent[1]{}%
  }%
  \providecommand\rotatebox[2]{#2}%
  \newcommand*\fsize{\dimexpr\f@size pt\relax}%
  \newcommand*\lineheight[1]{\fontsize{\fsize}{#1\fsize}\selectfont}%
  \ifx\svgwidth\undefined%
    \setlength{\unitlength}{483.57504056bp}%
    \ifx\svgscale\undefined%
      \relax%
    \else%
      \setlength{\unitlength}{\unitlength * \real{\svgscale}}%
    \fi%
  \else%
    \setlength{\unitlength}{\svgwidth}%
  \fi%
  \global\let\svgwidth\undefined%
  \global\let\svgscale\undefined%
  \makeatother%
  \begin{picture}(1,0.96141066)%
    \lineheight{1}%
    \setlength\tabcolsep{0pt}%
    \put(0,0){\includegraphics[width=\unitlength,page=1]{searchPoint.pdf}}%
    \put(0.74532242,0.36775241){\color[rgb]{0.78431373,0.21176471,0.21176471}\makebox(0,0)[lt]{\lineheight{1.25}\smash{\begin{tabular}[t]{l}$\mathbf{w}_{max}$\end{tabular}}}}%
    \put(0,0){\includegraphics[width=\unitlength,page=2]{searchPoint.pdf}}%
  \end{picture}%
\endgroup%

%% file: newSimplex.pdf_tex
\begingroup%
  \makeatletter%
  \providecommand\color[2][]{%
    \errmessage{(Inkscape) Color is used for the text in Inkscape, but the package 'color.sty' is not loaded}%
    \renewcommand\color[2][]{}%
  }%
  \providecommand\transparent[1]{%
    \errmessage{(Inkscape) Transparency is used (non-zero) for the text in Inkscape, but the package 'transparent.sty' is not loaded}%
    \renewcommand\transparent[1]{}%
  }%
  \providecommand\rotatebox[2]{#2}%
  \newcommand*\fsize{\dimexpr\f@size pt\relax}%
  \newcommand*\lineheight[1]{\fontsize{\fsize}{#1\fsize}\selectfont}%
  \ifx\svgwidth\undefined%
    \setlength{\unitlength}{483.57504056bp}%
    \ifx\svgscale\undefined%
      \relax%
    \else%
      \setlength{\unitlength}{\unitlength * \real{\svgscale}}%
    \fi%
  \else%
    \setlength{\unitlength}{\svgwidth}%
  \fi%
  \global\let\svgwidth\undefined%
  \global\let\svgscale\undefined%
  \makeatother%
  \begin{picture}(1,0.96141066)%
    \lineheight{1}%
    \setlength\tabcolsep{0pt}%
    \put(0,0){\includegraphics[width=\unitlength,page=1]{newSimplex.pdf}}%
  \end{picture}%
\endgroup%